\documentclass[sigconf, nonacm]{acmart}

\AtBeginDocument{%
  }

\usepackage{algorithm}
\usepackage{algpseudocode}
\usepackage{amsmath}
\usepackage{multirow}
\setcitestyle{numbers,comma}

\makeatletter
\newcommand{\doi}[1]{\url{https://doi.org/#1}}
\makeatother

\begin{document}

\title{GLACIER: A Multimodal Student-Teacher Foundation Model for Molecular Property Prediction}

\author{Emily Nguyen}
\authornote{Corresponding authors. This work has been accepted for publication in the Proceedings of the 32nd ACM SIGKDD Conference on Knowledge Discovery and Data Mining (KDD 2026), August 9--13, 2026, Jeju, South Korea.}
\orcid{0000-0003-4917-7336}
\affiliation{%
    \department{Department of Computer Science}
  \institution{University of Southern California}
  \city{Los Angeles}
  \state{California}
  \country{USA}
}
\email{emilyn98@usc.edu}

\author{Yongchan Hong}
\orcid{0009-0009-8866-1690}
\affiliation{%
 \department{Department of Quantitative and Computational Biology}
  \institution{University of Southern California}
  \city{Los Angeles}
  \state{California}
  \country{USA}
}
\email{hongyong@usc.edu}

\author{Harsh Toshniwal}
\orcid{0009-0008-2244-9497}
\affiliation{%
\department{Department of Computer Science}
  \institution{University of Southern California}
  \city{Los Angeles}
  \state{California}
  \country{USA}
}
\email{htoshniw@usc.edu}

\author{Yan Liu}
\orcid{0000-0002-7055-9518}
\affiliation{
  \institution{Amazon}
\department{Department of Computer Science}
  \institution{University of Southern California}
  \city{Los Angeles}
  \state{California}
  \country{USA}
}
\email{yanliu@cs.usc.edu}

\author{Andreas Luttens}
\authornotemark[1]
\orcid{0000-0003-2915-7901}
\affiliation{%
    \department{Department of Medical Biochemistry and Biophysics \\ Science for Life Laboratory}
    \institution{Karolinska Institutet}
  \city{Stockholm}
  \country{Sweden}
}
\email{andreas.luttens@ki.se}

\renewcommand{\shortauthors}{Nguyen et al.}

\begin{abstract}
Deep learning models facilitate the discovery of molecules with tailored properties among billions of candidate compounds. However, the computational burden to develop and deploy state-of-the-art models continuously increases, limiting their scalability. Most large-scale models are unimodal in nature and overlook the potential to leverage complementary molecular data modalities. To address these shortcomings, this paper introduces the Graph-Language Alignment for Chemical Inference and Exploration using Representations (GLACIER) model, a student-teacher framework that integrates molecular graphs, SMILES strings, and physicochemical descriptors to learn rich molecular embeddings. Our framework consists of three stages: (1) we pretrain three student encoders on $100,000$ drug-like molecules: a message-passing neural network for molecular graphs, a transformer-based encoder for SMILES strings, and a multilayer perceptron for physicochemical descriptors, (2) we fuse these student modalities using a novel Finsler geometry-aware module, and (3) distill complementary knowledge from large teacher models, including MiniMol and MolFormer, into a single lightweight model via contrastive learning. We demonstrate that GLACIER is a robust framework that delivers high predictive performance and computational efficiency in complex molecular property prediction tasks. Our code is publicly available at \href{https://github.com/eemokey/glacier}{https://github.com/eemokey/glacier}. 
\end{abstract}



\maketitle 

\begin{figure}[h]
  \includegraphics[width=\columnwidth]{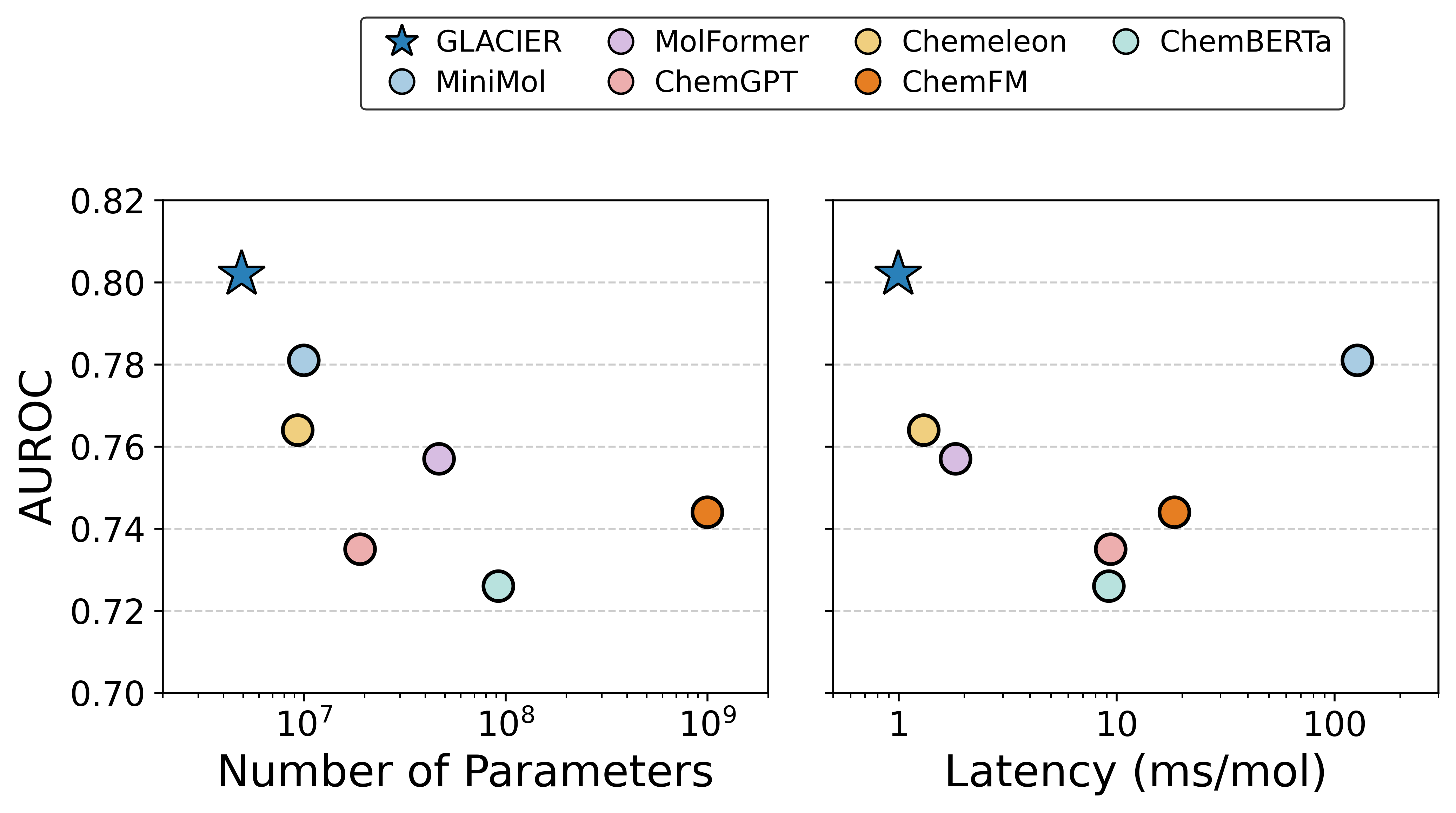}
  \caption{Model performance (AUROC) versus model parameter count (left) and model inference time per molecule (right).}
  \Description{Scatter plot with points for each model for efficiency vs AUROC.}
  \label{fig:teaser}
\end{figure}
\vspace{-1.5em}

\begin{figure*}[t]
  \centering
  \includegraphics[width=\linewidth]{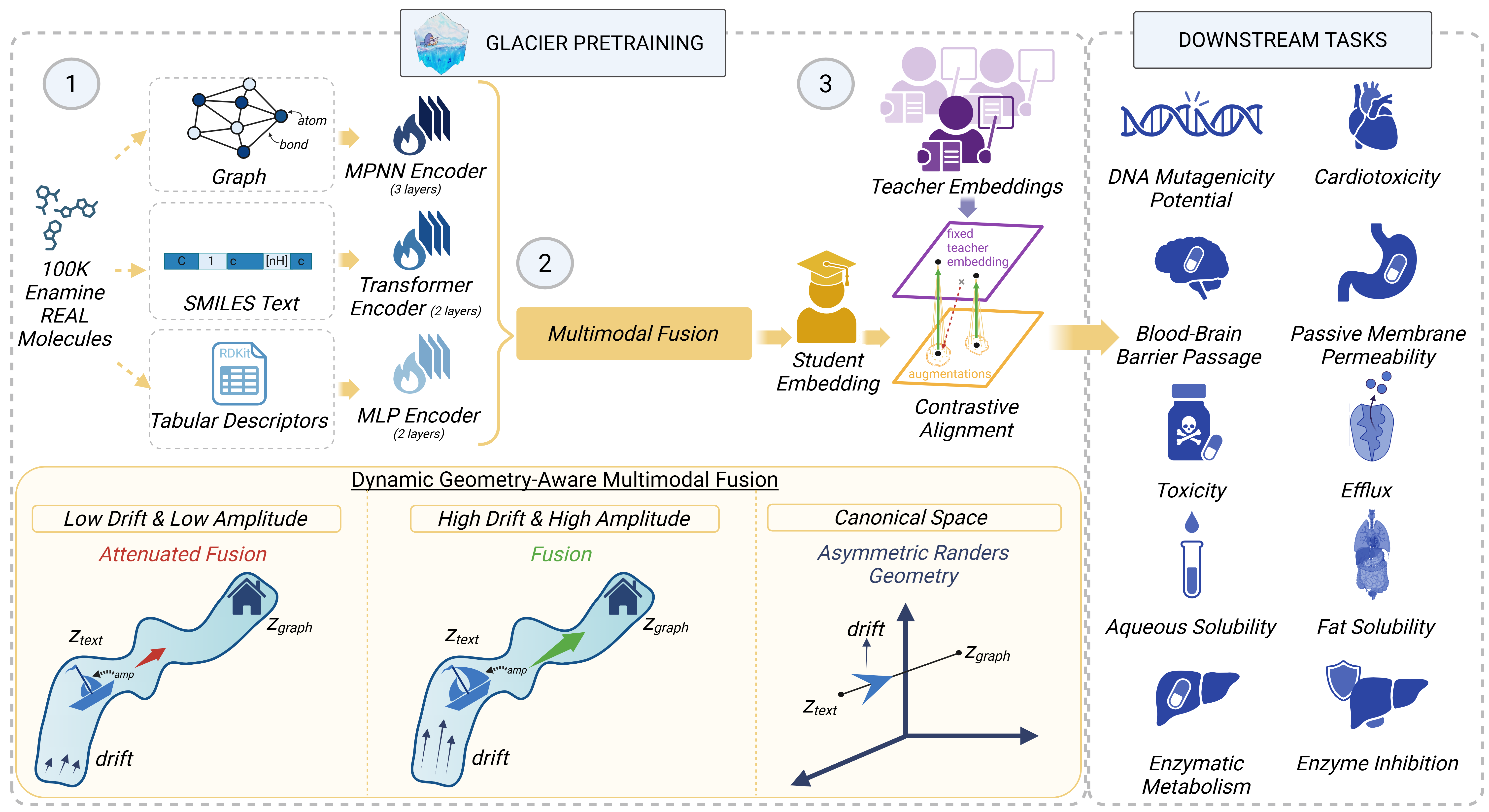} 
  \caption{Overview of the GLACIER framework: In Step 1, GLACIER is instantiated as a multimodal foundation model pretrained using 100,000 molecules sampled from the Enamine REAL database. The architecture processes each molecule across three modalities — molecular graphs, SMILES strings, and physicochemical descriptors — to capture a comprehensive molecular representation. In Step 2, the disparate modality representations obtained from Step 1 are integrated using a novel Finsler geometry-aware fusion mechanism that dynamically fuses graph, text, and tabular embeddings. In Step 3, the model is pretrained via teacher-to-student knowledge distillation using a contrastive objective that aligns the fused student embedding with fixed, large-scale teacher model embeddings. Finally, the model can be applied to downstream tasks.}
  \Description{Illustration of the GLACIER model architecture. Three student encoders using complementary molecular representations are geometrically fused. The resulting foundation model can be trained for downstream property prediction tasks relevant for drug discovery.}  
  \label{fig:overview_framework}
\end{figure*}

\section{Introduction}
Safe and efficacious drugs must exhibit a specific set of molecular properties, including potency against a drug target, selectivity, favorable pharmacokinetics and pharmacodynamics, and low toxicity \cite{hay2014adme}. Identifying molecules that satisfy these requirements is a lengthy and costly undertaking, often involving many cycles of design, synthesis, and experimental evaluation ~\cite{Wouters2020EstimatedRA}. To accelerate drug discovery, deep learning models are trained on chemical datasets to learn relationships between molecular structure and target properties, including biological activity and absorption, distribution, metabolism, excretion, and toxicity (ADMET) endpoints \cite{yang2019chemprop, Vamathevan2019ApplicationsOM, Swanson2024ADMETAIAM}. These models enable a more efficient prioritization of promising candidate compounds for downstream experimental evaluation \cite{stokes2020cell, krishnan2025cell, Luttens2025}.

Achieving this requires information-rich molecular representations and algorithms capable of mapping these representations to their corresponding properties. One promising approach is the use of chemical foundation models, which are first pretrained on large datasets to learn general chemical representations and then refined for specific downstream tasks using minimal additional data ~\cite{ChemFM, chithrananda2020chemberta, ross2022large}. To assess the predictive performance of these models, standardized benchmark datasets with experimentally measured properties are essential. Several public datasets, including Therapeutics Data Commons (TDC) and MoleculeNet, now serve as standard evaluation resources ~\cite{wu2018moleculenet, huang2021therapeutics}.

Many deep learning models achieve strong performance in molecular property prediction, but lack a more comprehensive chemical representation, struggle to generalize to different downstream tasks, or are very resource-intensive ~\cite{rong2020self, ross2022large, zhou2023uni}. This observation motivates the development of a lightweight model that leverages multiple molecular modalities for enhanced feature representation while supporting rapid deployment without compromising accuracy ~\cite{zhou2023uni, zeng2022deep, Klser2024MiniMolAP}. 

In this work, our contributions are as follows: 
\begin{enumerate}
    \item We propose Graph-Language Alignment for Chemical Inference and Exploration using Representations (GLACIER), a multimodal foundation model that learns unified molecular representations by distilling knowledge from state-of-the-art teacher models through contrastive pretraining on just $100,000$ drug-like molecules.
    \item We introduce a novel Finsler ~\cite{Cartan1933LesED, dages2025finsler} geometry-aware fusion mechanism for multimodal molecular representation learning, using a shared Randers space to dynamically align graph, SMILES \cite{Weininger1988SMILESAC}, and physicochemical descriptor embeddings and integrate complementary chemical information.
    \item We demonstrate that compact multimodal foundation models can rival and surpass substantially larger models, achieving state-of-the-art performance across molecular property prediction benchmarks while remaining lightweight and fast at inference. Our code and tutorials are publicly available at \href{https://github.com/eemokey/glacier}{https://github.com/eemokey/glacier}.
\end{enumerate}
\vspace{-1em} 

\section{Related work}
\subsection{Molecular representation learning} 
Existing molecular representation learning approaches can be broadly classified into three categories \cite{David2020MolecularRI, Praski2025Benchmarking}: (1) Graph neural network-based approaches: Methods such as GraphMVP ~\cite{liu2021pre} and GraphFP ~\cite{Luong2023FragmentbasedPA} leverage contrastive learning frameworks, while MiniMol ~\cite{Klser2024MiniMolAP} and Chemeleon ~\cite{Burns2025DescriptorbasedFM} provide structural insight, but are memory-intensive. (2) Transformer-based approaches: Models such as ChemBERTa \cite{chithrananda2020chemberta, Singh2025ChemBERTa3AO}, MolFormer \cite{ross2022large}, ChemGPT \cite{chemgpt}, ChemFM \cite{ChemFM}, MolBERT \cite{li2021mol}, and SimSon \cite{lee2025simson} improve the learning of global molecular representations with the self-attention mechanism, but they suffer from quadratic complexity \cite{vaswani2017attention}. (3) Hybrid-based approaches: Models that combine both graph-based and transformer-based approaches include GROVER \cite{rong2020self}, Uni-Mol \cite{zhou2023uni, ji2024uni}, and RMAT \cite{maziarka2024rmat}. However, these similarly suffer from high computational complexity that leads to longer training and inference times \cite{keles2023computational}. To tackle scalability challenges, knowledge distillation has emerged as a promising strategy, in which knowledge is transferred from large or ensemble teacher models to lightweight students ~\cite{ekstrom2023accelerating}. Despite the efficiency benefits of this paradigm, most molecular distillation methods are unimodal, and therefore overlook complementary insights present in different molecular representations. GLACIER distills the knowledge from large-scale chemical foundation models into a single lightweight model that integrates multimodal representations to overcome the challenges present in existing molecular property prediction approaches.

\subsection{Multimodal learning} 
Multimodal learning encompasses approaches that align or fuse data types for robust inference. The fusion of modalities such as molecular graphs, SMILES strings, and physicochemical descriptors remains challenging \cite{David2020MolecularRI}. Existing fusion methods include simple concatenation, cross-attention, and contrastive learning that align data into shared spaces \cite{radford2021learning}. Recent multimodal works include CL-FMAP~\cite{Zhou2025CLMFAPAC} (molecular graph, SMILES strings, Morgan fingerprints) and COATI~\cite{coati} (3D molecular conformers, SMILES), which leverage contrastive alignment across heterogeneous molecular representations to substantially improve model performance. Additional multimodal works include GIT-Mol \cite{Liu2023GITMolAM} (molecular graph, SMILES strings, images) and FineMolTex~\cite{finemoltex} (molecular graphs, textual descriptions) that merge modalities via cross-attention, further demonstrating the benefits of fusing structural and semantic molecular information. Following the precedent set by these works, we propose a framework that leverages geometrically fused representations of molecular graphs, SMILES strings, and physicochemical descriptors as an effective interface for distilling complementary knowledge from diverse teacher architectures into a single efficient model.

\section{The proposed approach}
In this section, we provide a detailed description of GLACIER's multimodal student-teacher distillation framework, as illustrated in Figure \ref{fig:overview_framework}\footnote{Created in BioRender. Nguyen, E. (2026) \href{https://BioRender.com/lg9qxrf}{https://BioRender.com/lg9qxrf}}. The architecture of the overall pipeline is presented Algorithms \ref{alg:finsler_fusion} and Algorithm \ref{alg:training_loop} in Appendix \ref{appendix:algs}.

\subsection{Step 1: Multimodal student architectures}

GLACIER integrates the information present in different modalities using encoders for each modality. The implementation in this work combines: (1) a graph encoder to extract information within molecular graphs; (2) a text encoder to extract information within SMILES strings, and (3) a tabular encoder to extract information from physicochemical descriptors.

\subsubsection{Graph encoder}
To capture topological information, we employ a Message Passing Neural Network (MPNN) ~\cite{Gilmer2017NeuralMPNN}. The molecule is represented as a directed graph $G=(V, E)$, where messages are passed iteratively between bonds and capture the local chemical environment. We perform $K=3$ message passing steps. To construct molecular embeddings $\mathbf{h}_{graph} \in \mathbb{R}^{300}$, we employ an attentive aggregation mechanism - a readout function that uses a learned weighted average to combine atom representations, enabling the model to dynamically prioritize chemically relevant substructures within a molecular graph. 
\begin{equation}
    \mathbf{h}_{graph} = \text{Readout}(\text{MPNN}(G)) 
\end{equation}

\subsubsection{Text encoder}
To capture sequential chemical patterns, the text encoder uses lightweight Transformer layers, consisting of $N=2$ layers with a hidden dimension of $d_{text}=128$ and eight attention heads. First, we process SMILES strings using a custom Byte-Pair Encoding (BPE) tokenizer trained on 100,000 randomly sampled molecules from the Enamine REAL database (65 billion, version 2024.07) \cite{enaminecite}. We optimize the vocabulary to a compact size of $V=8000$, prioritizing the learning of chemically semantic substructures over rare character combinations. The tokenizer maps a SMILES string $S$ to a fixed-length sequence of token indices $\mathbf{w} \in \mathbb{R}^L$, defined formally as:
\begin{equation}
     \mathbf{w} = \text{BPE}(S), \quad w_i \in \{0, \dots, V-1\}
 \end{equation}
where the sequence is padded to $L=512$ and includes special delimiters to define the molecular boundary of the attention mechanism. Then, we initialize the encoder input by summing learnable token embeddings with fixed sinusoidal positional encodings ($PE$) to retain sequence order information. The sequence is processed by the Transformer layers, and the output of the last hidden layer is pooled:
\begin{equation}
    \mathbf{h}_{text} = \text{Pool}(\text{Transformer}(\mathbf{w} + PE)) 
\end{equation}

\subsubsection{Tabular encoder}
Complementing the structural and sequential representations, we incorporate global physicochemical descriptors with a tabular encoder. The input consists of a feature vector $\mathbf{x}_{tab} \in \mathbb{R}^{217}$ computed by RDKit \cite{rdkitcite}. These descriptors include molecular properties such as molecular weight, logP, and the number of hydrogen bond donors and acceptors as described in the Table \ref{tab:rdkit_descriptors} in Appendix \ref{appendix:rdkit_desc}. The encoder is structured as an MLP, which yields the descriptor embedding:

\begin{equation}
    \mathbf{h}_{tab} = \text{MLP}(\mathbf{x}_{tab}) 
\end{equation}

\subsection{Step 2: Geometry-aware modality fusion}
After processing each modality through their encoders, we transform each through a dedicated projection head - implemented as a three-layer MLP - to map the representations into a shared latent space. We denote these projected embeddings as $\mathbf{z}_{graph}$, $\mathbf{z}_{text}$, and $\mathbf{z}_{tab}$ for molecular graph, text, and tabular embeddings, respectively. 

Using these modality embeddings, we propose a novel gated cross-attention fusion mechanism modeled on Finsler geometry for molecular representation learning, specifically adapting the asymmetric Randers metric \cite{randersPhysRev, dages2025finsler}. Unlike Riemannian metrics which measure distance isotropically, a Randers metric incorporates a directional drift vector field, effectively reducing the cost of transport in directions aligned with the drift. We adapt this to the semantic space by defining a drift vector $\boldsymbol{\omega}$ derived from the text embedding $\mathbf{z}_{text}$ where $\mathbf{v} = \text{MLP}_{drift}(\mathbf{z}_{text})$:

\begin{equation} 
\boldsymbol{\omega} = \frac{\mathbf{v}}{||\mathbf{v}||_2 + \epsilon} \cdot \tanh(||\mathbf{v}||_2)
\end{equation}

This creates a geometric bias where graph and tabular embeddings that align with the text's semantic direction are considered closer and thus more relevant.

Let $\mathbf{z}_{text}$ serve as the query and the set of complementary embeddings $S = \{\mathbf{z}_{graph}, \mathbf{z}_{tab}\}$ serve as the keys. The asymmetric Randers distance $d$ is defined as the combination of the Euclidean distance and the projection onto the drift vector:
\begin{equation}
    d(\mathbf{z}_{text}, \mathbf{k}) = \|\mathbf{k} - \mathbf{z}_{text}\|_2 + \langle \mathbf{k} - \mathbf{z}_{text}, \boldsymbol{\omega} \rangle  
\end{equation}

An attention correction vector $\mathbf{c}$ is computed via softmax over these negative distances. To balance the integration of this correction, we adopt a text-contextualized approach that dynamically adjusts the importance of the modalities. GLACIER learns a scalar amplitude $\alpha$, which modulates a sigmoid gate \cite{qiu2025gated} based on the minimum geometric distance:

\begin{equation}
    \gamma = \alpha(\mathbf{z}_{text}) \cdot \sigma\left(-\min_{\mathbf{k} \in S} d(\mathbf{z}_{text}, \mathbf{k}) \cdot \lambda\right)
\end{equation}

Here, the learnable parameters serve three geometric roles: the weights of $\text{MLP}_{drift}$ learn the optimal semantic direction for fusion; $\text{MLP}_{amp}$ learns the confidence magnitude $\alpha$, allowing the model to determine how much additional information to accept; and the scalar $\lambda$ (gate sensitivity) learns the curvature of the gating function, controlling how strictly geometric misalignment is penalized. The text embedding is refined as $\mathbf{\hat{z}}_{text} = \mathbf{z}_{text} + \gamma \mathbf{c}$, and the final fused embedding $\mathbf{h}_{fused}$ is obtained by concatenating the refined text embeddings with the molecular graph and tabular embeddings.

\subsection{Step 3: Student-teacher knowledge distillation}

To distill knowledge from large-scale models into our lightweight architecture, we align the fused student embeddings with one or multiple fixed teacher embeddings. We investigate distillation from two high-performing teachers, each representing a different model family: (1) a graph-based teacher, MiniMol and (2) a transformer-based teacher, MolFormer. 

\subsubsection{Projection distillation layers}

We utilize diverse sets of $K$ teacher models, each providing precomputed, fixed embeddings $\mathbf{t}_k$ of varying dimensionality and architectural origin. To align the student with the teacher, we employ independent teacher projections $\{P_k\}_{k=1}^K$, each consisting of a two-layer MLP to project the embeddings of the teacher into the shared dimension $d_{shared}=512$. Simultaneously, the fused student embedding $\mathbf{h}_{fused}$ has its own projector layer ($P_S$). This standard module decouples the geometric fusion space from the direct gradients of the alignment loss. The final embeddings for alignment are the following:
\begin{equation}
    \mathbf{z}_{S} = P_S(\mathbf{h}_{fused}), \quad \mathbf{z}_{T}^{(k)} = P_k(\text{stop\_grad}(\mathbf{t}_k))
\end{equation}

\subsubsection{Distillation objective}
Standard multi-teacher distillation often treats all teachers equally, which is suboptimal when teachers have varying expertise. To address this, we introduce a dynamic multi-teacher InfoNCE loss that allows the student to dynamically adjust the contribution of each teacher \cite{oord2018representation}.
We employ an internal contribution head, $T(\cdot)$, a two-layer MLP that predicts a contribution score $\tau_k \in [\epsilon, 1.0]$ for each teacher based on the current embedding of the student $\mathbf{z}_{S}$. To prevent the model from completely ignoring difficult teachers, we enforce a minimum contribution floor $\epsilon=0.1$:
\begin{equation}
    \tau_k = \sigma(\text{MLP}_{contribution}(\mathbf{z}_{S})) \cdot (1 - \epsilon) + \epsilon
\end{equation}

The total loss is calculated as the weighted sum of the InfoNCE loss $\mathcal{L}_{NCE}$ for each teacher, regularized by a logarithmic term to prevent collapse:
\begin{equation}
    \mathcal{L} = \sum_{k=1}^K \left( \tau_k \cdot \mathcal{L}_{NCE}(\mathbf{z}_{S}, \mathbf{z}_{T}^{(k)}) - \log(\tau_k) \right)
\end{equation}
Thus, GLACIER can jointly learn and distill knowledge from multiple teachers. 

\section{Experiments}
\begin{table*}[t]
\caption{AUROC scores for molecular property prediction on TDC and MoleculeNet. The best results are marked in \textbf{bold}, and the second-best results are \underline{underlined}. $\uparrow$: the higher the better. Values represent means and their standard deviations from three independent runs.}
\label{tab:main_results}
\centering
\small
\setlength{\tabcolsep}{1.5 pt}
\begin{tabular}{lcccccccccc}
\toprule
Method & AMES $\uparrow$ & BBB  $\uparrow$ & Pgp $\uparrow$ & E-Sub $\uparrow$ & E-Inh $\uparrow$ & hERG $\uparrow$  &PAMPA $\uparrow$ & Tox21 $\uparrow$ & ToxCast $\uparrow$ &  Avg $\uparrow$  
\\ \midrule

\multicolumn{11}{l}{\textit{Graph Neural Networks}} \\

MiniMol & 0.818\textsubscript{$\pm$ 0.004} & 0.813\textsubscript{$\pm$ 0.036} & 0.879\textsubscript{$\pm$ 0.030} & 0.787\textsubscript{$\pm$ 0.121} & \textbf{0.857\textsubscript{$\pm$ 0.007}} & 0.801\textsubscript{$\pm$ 0.026} & 0.623\textsubscript{$\pm$ 0.066} & 0.770\textsubscript{$\pm$ 0.009} & \textbf{0.680\textsubscript{$\pm$ 0.011}} & 0.781 \\

Chemeleon & 0.760\textsubscript{$\pm$ 0.038} & 0.849\textsubscript{$\pm$ 0.024} & 0.784\textsubscript{$\pm$ 0.075} & \textbf{0.819\textsubscript{$\pm$ 0.008}} & 0.634\textsubscript{$\pm$ 0.041} & \underline{0.884\textsubscript{$\pm$ 0.033}} & \textbf{0.784\textsubscript{$\pm$ 0.014}} & 0.728\textsubscript{$\pm$ 0.058} & 0.636\textsubscript{$\pm$ 0.021} & 0.764 \\

\midrule
\multicolumn{11}{l}{\textit{Text Transformers}} \\

ChemBERTa-100M & 0.755\textsubscript{$\pm$ 0.045} & 0.805\textsubscript{$\pm$ 0.022} & 0.899\textsubscript{$\pm$ 0.029} & 0.707\textsubscript{$\pm$ 0.069} & 0.769\textsubscript{$\pm$ 0.014} & 0.742\textsubscript{$\pm$ 0.009} & 0.588\textsubscript{$\pm$ 0.064} & 0.679\textsubscript{$\pm$ 0.034} & 0.593\textsubscript{$\pm$ 0.019} & 0.726 \\

MolFormer & 0.795\textsubscript{$\pm$ 0.028} & 0.865\textsubscript{$\pm$ 0.017} & 0.881\textsubscript{$\pm$ 0.042} & 0.759\textsubscript{$\pm$ 0.037} & 0.804\textsubscript{$\pm$ 0.011} & 0.778\textsubscript{$\pm$ 0.016} & 0.605\textsubscript{$\pm$ 0.040} & 0.714\textsubscript{$\pm$ 0.040} & 0.615\textsubscript{$\pm$ 0.009} & 0.757 \\

ChemGPT-19M & 0.740\textsubscript{$\pm$ 0.035} & 0.817\textsubscript{$\pm$ 0.050} & 0.885\textsubscript{$\pm$ 0.019} & 0.720\textsubscript{$\pm$ 0.016} & 0.777\textsubscript{$\pm$ 0.005} & 0.722\textsubscript{$\pm$ 0.017} & 0.631\textsubscript{$\pm$ 0.031} & 0.708\textsubscript{$\pm$ 0.030} & 0.612\textsubscript{$\pm$ 0.004} & 0.735 \\

ChemFM-1B & 0.737\textsubscript{$\pm$ 0.021} & 0.872\textsubscript{$\pm$ 0.012} & 0.883\textsubscript{$\pm$ 0.045} & 0.711\textsubscript{$\pm$ 0.045} & 0.758\textsubscript{$\pm$ 0.015} & 0.747\textsubscript{$\pm$ 0.020} & 0.668\textsubscript{$\pm$ 0.016} & 0.703\textsubscript{$\pm$ 0.016} & 0.612\textsubscript{$\pm$ 0.013} & 0.744 \\ \midrule

\multicolumn{11}{l}{\textit{Hybrid Models}} \\
RMAT-4M & 0.797\textsubscript{$\pm$ 0.031} & 0.844\textsubscript{$\pm$ 0.025} & 0.897\textsubscript{$\pm$ 0.031} & 0.701\textsubscript{$\pm$ 0.074} & 0.806\textsubscript{$\pm$ 0.003} & 0.798\textsubscript{$\pm$ 0.027} & 0.683\textsubscript{$\pm$ 0.036} & 0.755\textsubscript{$\pm$ 0.027} & 0.640\textsubscript{$\pm$ 0.009} & 0.769 \\

COATI & 0.799\textsubscript{$\pm$ 0.013} & 0.815\textsubscript{$\pm$ 0.041} & 0.752\textsubscript{$\pm$ 0.102} & \underline{0.814\textsubscript{$\pm$ 0.005}} & 0.685\textsubscript{$\pm$ 0.070} & \textbf{0.902\textsubscript{$\pm$ 0.038}} & \underline{0.743\textsubscript{$\pm$ 0.025}} & 0.740\textsubscript{$\pm$ 0.016} & 0.633\textsubscript{$\pm$ 0.007} & 0.765 \\

CL-MFAP & 0.742\textsubscript{$\pm$ 0.031} & 0.801\textsubscript{$\pm$ 0.028} & 0.862\textsubscript{$\pm$ 0.041} & 0.748\textsubscript{$\pm$ 0.034} & 0.771\textsubscript{$\pm$ 0.019} & 0.759\textsubscript{$\pm$ 0.022} & 0.612\textsubscript{$\pm$ 0.047} & 0.701\textsubscript{$\pm$ 0.033} & 0.608\textsubscript{$\pm$ 0.015} & 0.734 \\

GIT-Mol & 0.758\textsubscript{$\pm$ 0.006} & 0.813\textsubscript{$\pm$ 0.019} & 0.864\textsubscript{$\pm$ 0.034} & 0.731\textsubscript{$\pm$ 0.047} & 0.771\textsubscript{$\pm$ 0.015} & 0.745\textsubscript{$\pm$ 0.014} & 0.666\textsubscript{$\pm$ 0.011} & 0.571\textsubscript{$\pm$ 0.026} & 0.598\textsubscript{$\pm$ 0.008} & 0.724 \\

\midrule

\multicolumn{11}{l}{\textit{OURS}} \\

GLACIER (MolFormer) & 0.809\textsubscript{$\pm$ 0.027} & 0.861\textsubscript{$\pm$ 0.028} & \underline{0.908\textsubscript{$\pm$ 0.031}} & 0.785\textsubscript{$\pm$ 0.041} & 0.844\textsubscript{$\pm$ 0.011} & 0.773\textsubscript{$\pm$ 0.018} & 0.638\textsubscript{$\pm$ 0.087} & 0.758\textsubscript{$\pm$ 0.034} & 0.644\textsubscript{$\pm$ 0.011} & 0.780 \\

GLACIER (MiniMol)  & \textbf{0.828\textsubscript{$\pm$ 0.018}} & \textbf{0.895\textsubscript{$\pm$ 0.027}} & 0.900\textsubscript{$\pm$ 0.029} & 0.769\textsubscript{$\pm$ 0.121} & \underline{0.856\textsubscript{$\pm$ 0.011}} & 0.803\textsubscript{$\pm$ 0.026} & 0.687\textsubscript{$\pm$ 0.079} & \textbf{0.786\textsubscript{$\pm$ 0.034}} & \underline{0.668\textsubscript{$\pm$ 0.008}} & \textbf{0.799} \\

GLACIER (Mi-Mo) & \underline{0.824\textsubscript{$\pm$ 0.022}} & \underline{0.883\textsubscript{$\pm$ 0.031}} & \textbf{0.915\textsubscript{$\pm$ 0.027}} & 0.768\textsubscript{$\pm$ 0.076} & 0.848\textsubscript{$\pm$ 0.009} & 0.783\textsubscript{$\pm$ 0.021} & 0.657\textsubscript{$\pm$ 0.089} & \underline{0.783\textsubscript{$\pm$ 0.038}} & 0.663\textsubscript{$\pm$ 0.012} & \underline{0.792} \\

\bottomrule
\end{tabular}
\end{table*}

\subsection{Pretraining GLACIER}  
To construct the pretraining corpus, we randomly sampled $100,000$ molecules from the Enamine REAL database (65 billion molecules, version 2024.07) \cite{enaminecite}, chosen for its extensive collection of synthetically accessible, drug-like compounds \cite{Grygorenko2020GeneratingMC}. An assessment of potential overlap between the pretraining corpus and downstream benchmarks is provided in Figure \ref{fig:tanimoto_sim} in Appendix \ref{appendix:pretraining_sim}. The ChemAxon Extended SMILES (CXSMILES) annotations \cite{cxsmilescite} were removed, retaining only the canonical SMILES strings. These standardized molecules were then used to generate the three pretraining modalities employed by GLACIER: molecular graphs, SMILES strings, and physicochemical descriptors. 

To improve representation learning during pretraining, we employed dynamic SMILES augmentation by generating a randomized valid SMILES string for each molecule at every epoch. This approach exploits the fact that a single molecular graph can be represented by multiple equivalent SMILES strings depending on the choice of starting atom and graph traversal order. By exposing the model to diverse textual realizations of the same underlying structure, this stochasticity reduces the reliance on specific syntactic patterns and encourages the learning of chemically invariant representations \cite{Bjerrum2017SMILESEA}. As these alternative SMILES representations are generated on-the-fly, they increase representation diversity without requiring additional molecular data or substantial computational overhead.

For knowledge distillation, we used MiniMol and MolFormer as teacher models. The teacher embeddings were extracted once and reused throughout pretraining, making the distillation process computationally efficient. GLACIER was pretrained for 250 epochs in 5.67 hours on a single NVIDIA RTX 4080 GPU, highlighting the modest computational requirements of the framework. Additional implementation and hardware details are provided in Tables \ref{tab:model_config} and \ref{tab:training_dynamics} in Appendix \ref{appendix:implementation_details}.

\subsection{Molecular benchmark datasets}
We evaluated GLACIER's performance on 11 molecular property prediction tasks taken from two main benchmarks relevant for drug discovery: TDC \cite{huang2021therapeutics} and MoleculeNet \cite{wu2018moleculenet}. These datasets span two broad property prediction scenarios: (1) Molecular classification datasets: AMES, BBB , Pgp, E-Sub, E-Inh, hERG, PAMPA, Tox21, and ToxCast; (2) Molecular regression datasets: ESOL and LIPO. These datasets vary in both the number of classes, from 2 to 617 classes, and in the total number of samples, from 664  to 13,192 molecules. This allows us to verify our distillation method for a broad range of configurations and ensure its applicability. A numerical overview of the datasets and descriptions of their corresponding tasks are provided in Tables \ref{tab:dataset_stats} and  \ref{tab:benchmark_description} in Appendix \ref{appendix:dataset_details}.

\subsection{Baselines} 
We compared GLACIER against a range of recent baselines spanning diverse methodologies, including graph neural network–based models (MiniMol~\cite{Klser2024MiniMolAP} and Chemeleon~\cite{Burns2025DescriptorbasedFM}) and text-based transformer models (ChemBERTa~\cite{Singh2025ChemBERTa3AO},  MolFormer~\cite{ross2022large}, ChemGPT~\cite{chemgpt}, and ChemFM-1B~\cite{ChemFM}). We also evaluated hybrid models, including RMAT \cite{maziarka2024rmat}, COATI~\cite{coati}, CL-FMAP~\cite{Zhou2025CLMFAPAC}, and GIT-Mol \cite{Liu2023GITMolAM}.

We organize our experiments around the following research questions (RQs): 
\begin{itemize}
\item \textbf{RQ1}: Does GLACIER perform well on downstream tasks? 
\item \textbf{RQ2}: Does GLACIER outperform its baseline teachers?
\item \textbf{RQ3}: Does GLACIER produce interpretable embeddings?
\item \textbf{RQ4}: Does GLACIER have an optimal fusion mechanism, modality composition, and pretraining scale?
\end{itemize}

\begin{figure*}[t]
  \centering
  \includegraphics[width=\linewidth]{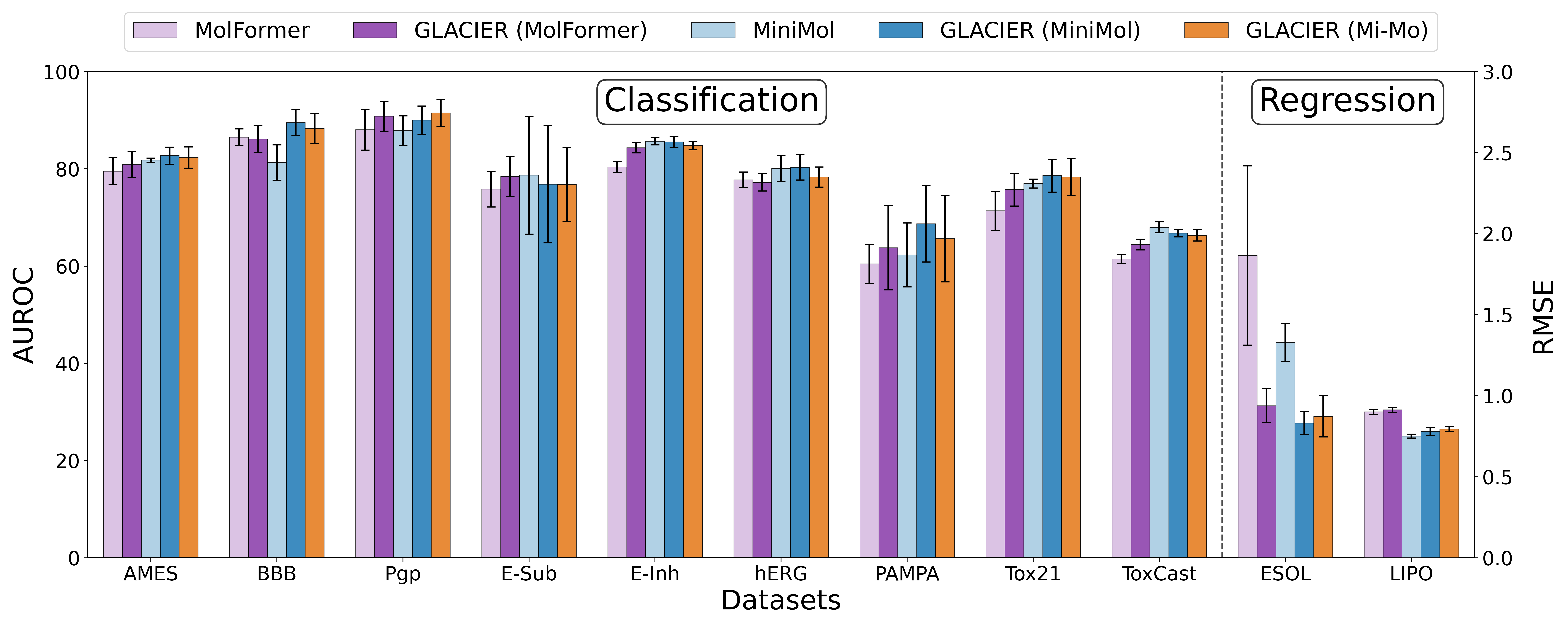}
  \caption{Performance comparison across molecular property prediction tasks. Muted colors represent teacher baselines, while saturated colors represent their respective student version in a GLACIER-Finsler distillation framework (MolFormer in purple, MiniMol in blue, Mi-Mo in orange). Nine datasets are used for classification tasks, the remaining two are regression tasks. Error bars correspond to the standard deviation of the mean across three independent runs.}
  \Description{Plot of student-teacher performance comparisons where the student outperforms the teacher on most classification and regression. } 
  \label{fig:teacher-student}
\end{figure*}

\begin{table}[h]
\caption{RMSE scores for molecular property prediction on MoleculeNet. The best results are marked in \textbf{bold}, and the second-best results are \underline{underlined}. $\downarrow$: the lower the better. Values represent means and their standard deviations from three independent runs.}
\label{tab:main_reg_results}
\centering
\small
\setlength{\tabcolsep}{2.5pt} 
\begin{tabular}{lcccc}
\toprule
Method & ESOL $\downarrow$ & LIPO $\downarrow$ &  Avg $\downarrow$ \\ 
\midrule
\multicolumn{4}{l}{\textit{Graph Neural Networks}} \\
MiniMol & 1.328\textsubscript{$\pm$ 0.117} & \textbf{0.751\textsubscript{$\pm$ 0.012}} & 1.040 \\
Chemeleon & 2.417\textsubscript{$\pm$ 1.091} & 1.416\textsubscript{$\pm$ 0.052} & 1.916 \\

\midrule

\multicolumn{4}{l}{\textit{Text Transformers}} \\
ChemBERTa-100M & 1.808\textsubscript{$\pm$ 0.053} & 1.069\textsubscript{$\pm$ 0.033} & 1.439\\
MolFormer & 1.866\textsubscript{$\pm$ 0.553} & 0.901\textsubscript{$\pm$ 0.016} & 1.383 \\
ChemGPT-19M & 1.354\textsubscript{$\pm$ 0.355} & 1.020\textsubscript{$\pm$ 0.017} & 1.187 \\
ChemFM-1B & 1.762\textsubscript{$\pm$ 0.316} & 1.406\textsubscript{$\pm$ 0.008} & 1.584 \\ \midrule

\multicolumn{4}{l}{\textit{Multimodal Models}} \\
RMAT-4M & 1.305\textsubscript{$\pm$ 0.080} & 1.029\textsubscript{$\pm$ 0.070} & 1.167 \\
COATI & 1.038\textsubscript{$\pm$ 0.122} & 0.924\textsubscript{$\pm$ 0.058} & 0.981 \\
CL-MFAP & 1.134\textsubscript{$\pm$ 0.184} & 1.059\textsubscript{$\pm$ 0.061} & 1.096 \\
GIT-Mol & 2.336\textsubscript{$\pm$ 1.088} & 1.109\textsubscript{$\pm$ 0.030} & 1.722 \\

\midrule

\multicolumn{4}{l}{\textit{OURS}} \\
GLACIER (MolFormer) & 0.939\textsubscript{$\pm$ 0.105} & 0.913\textsubscript{$\pm$ 0.015} & 0.926 \\
GLACIER (MiniMol) & \textbf{0.831\textsubscript{$\pm$ 0.071}} & \underline{0.780\textsubscript{$\pm$ 0.025}} & \textbf{0.806} \\
GLACIER (Mi-Mo) & \underline{0.873\textsubscript{$\pm$ 0.126}} & 0.795\textsubscript{$\pm$ 0.015} & \underline{0.834} \\
\bottomrule
\end{tabular}
\end{table}

\subsection{RQ1: Downstream property predictions}
We evaluated GLACIER models built using three different teacher configurations: MolFormer as a single teacher, MiniMol as a single teacher, and the combination of MiniMol and MolFormer as teachers (Mi-Mo). A detailed explanation on the choice of teachers is provided in Appendix \ref{appendix:teacher_selection}. For downstream evaluation, we conducted downstream fingerprinting, which is more computationally efficient and practical compared to end-to-end finetuning \cite{jiang2025trident, Klser2024MiniMolAP, Praski2025Benchmarking}. Specifically, we extracted frozen embeddings of the final layer of GLACIER for molecules in a given downstream task. These embeddings were used to train a small task head (logistic regression) to make task-specific predictions. Following the benchmarks of TDC \cite{huang2021therapeutics} and MoleculeNet \cite{wu2018moleculenet}, we used AUROC (Area Under Receiver Operating Characteristic Curve) as an evaluation metric for classification tasks and RMSE (Root Mean Squared Error) for regression tasks. The molecules in each benchmark dataset underwent a standardization process using RDKit \cite{rdkitcite}. This includes the removal of salts, neutralization of charges, canonicalization of SMILES strings, and the removal of duplicates. We then used an 80/10/10 scaffold split for training, validation, and testing to evaluate generalization to unseen chemical scaffolds \cite{wu2018moleculenet}. More details on task evaluations are provided in Appendix \ref{appendix:evaluation_task}.
 
Although the results reported in Tables \ref{tab:main_results} and \ref{tab:main_reg_results} indicate that molecular property prediction remains challenging, two observations arise from our analyzes: First, GLACIER on average outperforms other models on the classification and regression benchmarks. This suggests that geometry-aware fusion coupled with contrastive distillation contributes to a latent space that successfully captures relevant molecular features, leading to a model that generalizes well to various property prediction tasks. Second, we observe that compact models can outperform substantially larger foundation models, indicating that gains in predictive performance cannot be achieved through parameter scaling alone.

\begin{figure}[h]
  \includegraphics[width=\columnwidth]{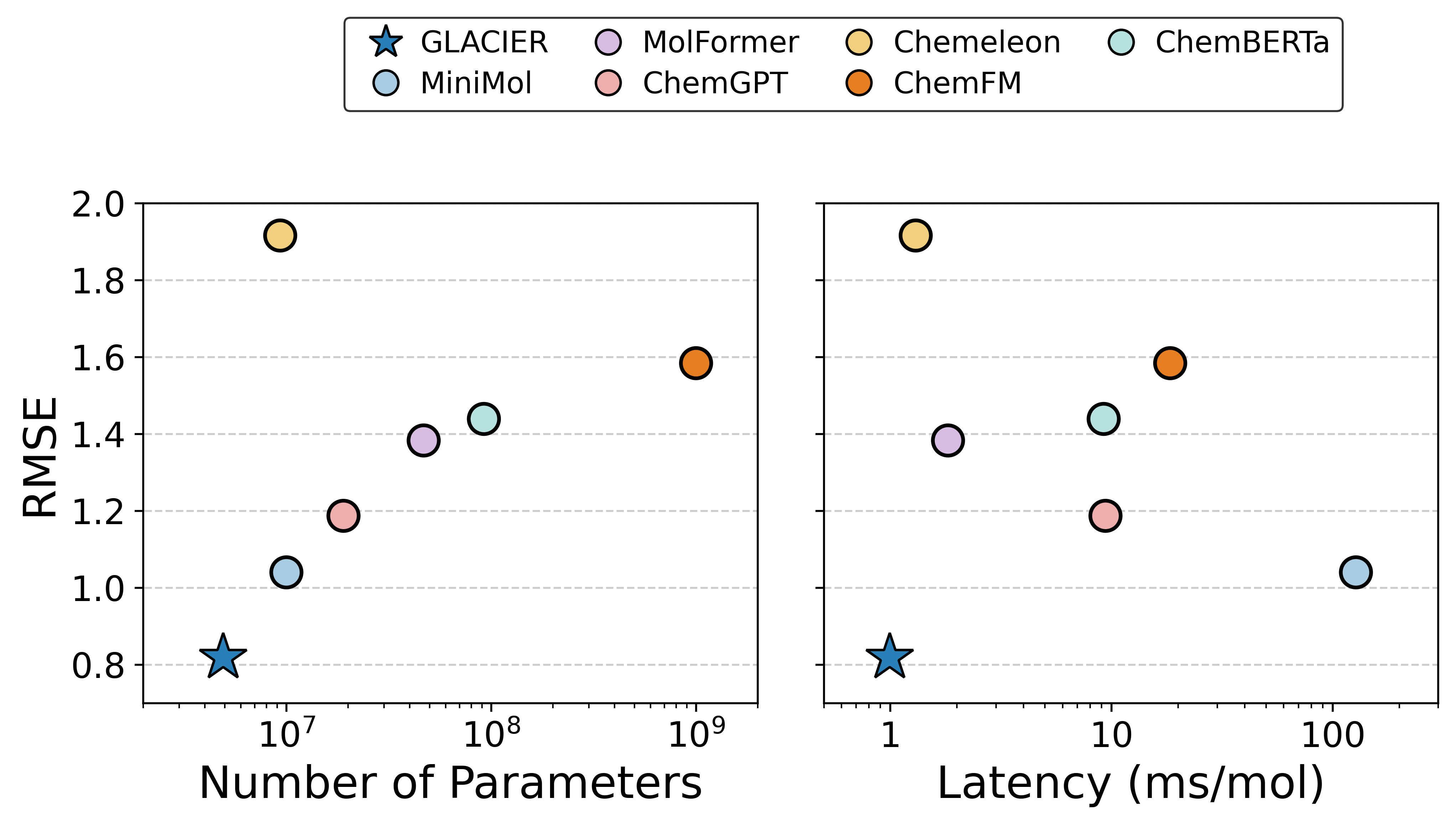}
  \caption{Model performance (RMSE) versus model parameter count (left) and model inference time per molecule (right).}
  \Description{Scatter plot with points for each model for efficiency vs RMSE.}
  \label{fig:teaser_RMSE}
\end{figure}

\begin{figure}[t]  
    \centering
    \includegraphics[width=\columnwidth]{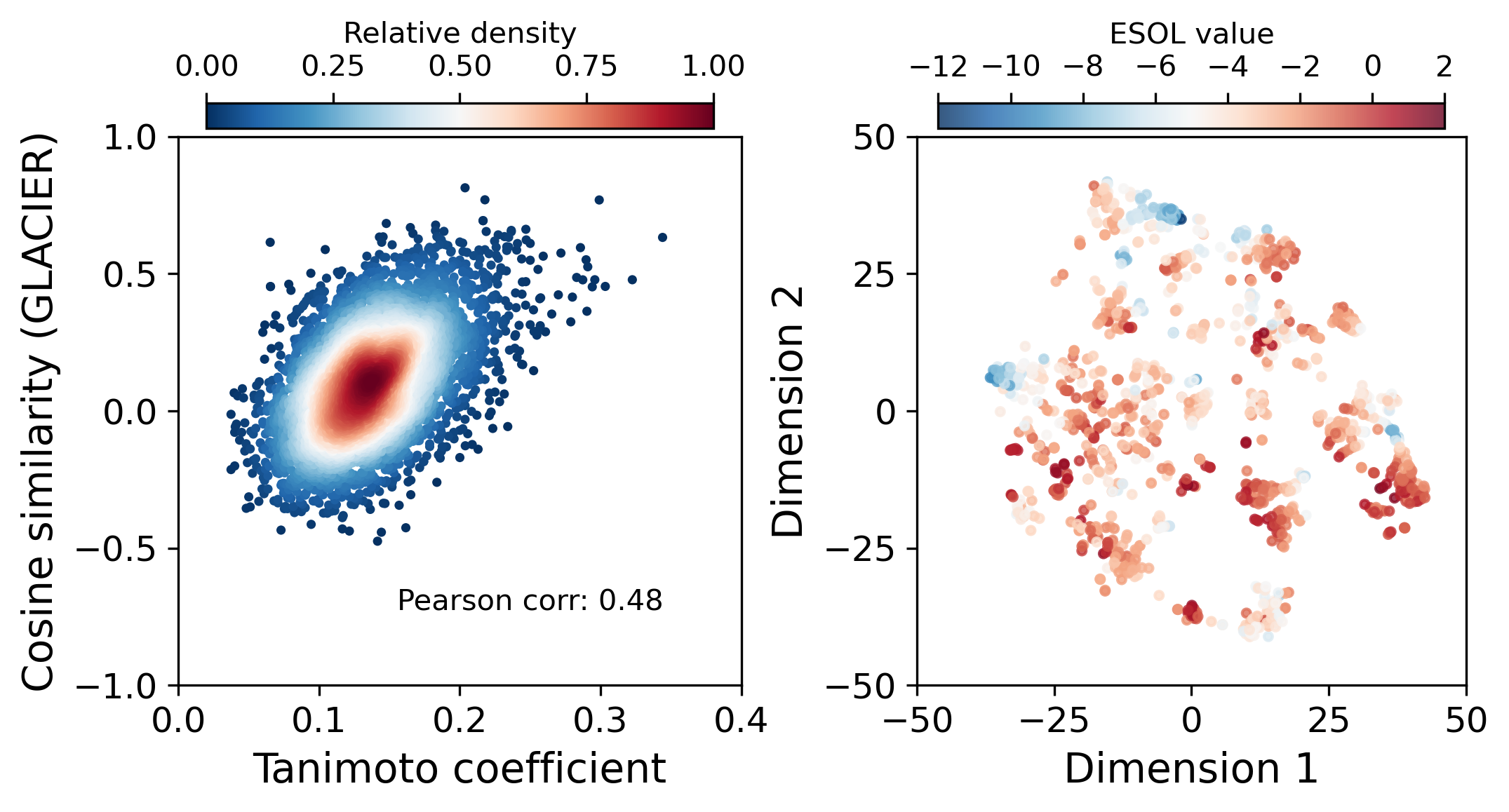}
    \caption{(Left) Two-dimensional density-normalized scatter plot assessing the alignment between cosine similarity in the GLACIER embedding space and Tanimoto coefficients for corresponding molecules. (Right) Two-dimensional t-SNE projection of 512-dimensional GLACIER embeddings for molecules from the MoleculeNet ESOL dataset, illustrating the structure of the learned representation.}
    \Description{Interpretability plots of the GLACIER embedding space. The left plot shows a scatter plot with a trend. The right plot shows a t-SNE visualization. }
    \label{fig:embedding_vis}
\end{figure}

\begin{table*}[t]
\caption{Ablation study comparing Concatenation vs Finsler fusion using AUROC scores on TDC and MoleculeNet. Best results are marked in \textbf{bold}. $\uparrow$: higher is better. Second-best results are \underline{underlined}. Values represent means and their standard deviations from three independent runs.}
\label{tab:cls_ablation}
\centering
\small
\setlength{\tabcolsep}{1 pt} 
\begin{tabular}{lcccccccccc}
\toprule
Method & AMES $\uparrow$ & BBB $\uparrow$ & Pgp $\uparrow$ & E-Sub $\uparrow$ & E-Inh $\uparrow$ & hERG $\uparrow$ & PAMPA $\uparrow$ & Tox21 $\uparrow$ & ToxCast $\uparrow$ & Avg $\uparrow$ \\ \midrule

\multicolumn{11}{l}{\textit{MolFormer Teacher}} \\
GLACIER (Concat) & \textbf{0.817\textsubscript{$\pm$ 0.021}} & 0.844\textsubscript{$\pm$ 0.028} & \textbf{0.914\textsubscript{$\pm$ 0.030}} & 0.766\textsubscript{$\pm$ 0.047} & \underline{0.839\textsubscript{$\pm$ 0.008}} & \underline{0.790\textsubscript{$\pm$ 0.023}} & \textbf{0.687\textsubscript{$\pm$ 0.076}} & \textbf{0.781\textsubscript{$\pm$ 0.033}} & \textbf{0.649\textsubscript{$\pm$ 0.013}} & \textbf{0.788} \\ 

GLACIER (Attention) & \underline{0.814\textsubscript{$\pm$ 0.042}} & \underline{0.850\textsubscript{$\pm$ 0.030}} & 0.904\textsubscript{$\pm$ 0.022} & \textbf{0.813\textsubscript{$\pm$ 0.104}} & 0.830\textsubscript{$\pm$ 0.012} & \textbf{0.795\textsubscript{$\pm$ 0.027}} & \underline{0.672\textsubscript{$\pm$ 0.061}} & 0.748\textsubscript{$\pm$ 0.042} & 0.634\textsubscript{$\pm$ 0.009} & \underline{0.784} \\

GLACIER (Finsler) & 0.809\textsubscript{$\pm$ 0.027} & \textbf{0.861\textsubscript{$\pm$ 0.028}} & \underline{0.908\textsubscript{$\pm$ 0.031}} & \underline{0.785\textsubscript{$\pm$ 0.041}} & \textbf{0.844\textsubscript{$\pm$ 0.011}} & 0.773\textsubscript{$\pm$ 0.018} & 0.638\textsubscript{$\pm$ 0.087} & \underline{0.758\textsubscript{$\pm$ 0.034}} & \underline{0.644\textsubscript{$\pm$ 0.011}} & 0.780 \\

\midrule

\multicolumn{11}{l}{\textit{MiniMol Teacher}} \\
GLACIER (Concat) & 0.813\textsubscript{$\pm$ 0.020} & \underline{0.884\textsubscript{$\pm$ 0.027}} & \textbf{0.920\textsubscript{$\pm$ 0.023}} & \underline{0.753\textsubscript{$\pm$ 0.108}} & \textbf{0.856\textsubscript{$\pm$ 0.009}} & \underline{0.789\textsubscript{$\pm$ 0.028}} & \textbf{0.699\textsubscript{$\pm$ 0.072}} & \textbf{0.786\textsubscript{$\pm$ 0.028}} & \textbf{0.668\textsubscript{$\pm$ 0.012}} & \underline{0.796} \\

GLACIER (Attention) & \underline{0.817\textsubscript{$\pm$ 0.020}} & 0.869\textsubscript{$\pm$ 0.019} & 0.894\textsubscript{$\pm$ 0.032} & 0.707\textsubscript{$\pm$ 0.048} & \underline{0.847\textsubscript{$\pm$ 0.014}} & \textbf{0.803\textsubscript{$\pm$ 0.027}} & \textbf{0.699\textsubscript{$\pm$ 0.079}} & \underline{0.757\textsubscript{$\pm$ 0.037}} & \underline{0.652\textsubscript{$\pm$ 0.012}} & 0.783 \\

GLACIER (Finsler) & \textbf{0.828\textsubscript{$\pm$ 0.018}} & \textbf{0.895\textsubscript{$\pm$ 0.027}} & \underline{0.900\textsubscript{$\pm$ 0.029}} & \textbf{0.769\textsubscript{$\pm$ 0.121}} & \textbf{0.856\textsubscript{$\pm$ 0.011}} & \textbf{0.803\textsubscript{$\pm$ 0.026}} & \underline{0.687\textsubscript{$\pm$ 0.079}} & \textbf{0.786\textsubscript{$\pm$ 0.034}} & \textbf{0.668\textsubscript{$\pm$ 0.008}} & \textbf{0.799} \\

\bottomrule
\end{tabular}
\end{table*}

\subsection{RQ2: Distillation efficacy}
We evaluated the performance of GLACIER models in relation to their respective teacher models across 11 benchmark datasets. In particular, we considered the graph-based model MiniMol and the transformer-based model MolFormer, comparing both single-teacher and dual-teacher distillation strategies. The single-teacher variants used either MiniMol or MolFormer alone, whereas the dual-teacher variant leveraged both models simultaneously during pretraining. Figure \ref{fig:teacher-student} shows the performance of our student GLACIER models compared to their baseline teachers. Two main observations emerge. First, GLACIER consistently achieves comparable or superior performance, outperforming its respective teacher baselines across the majority of benchmarks. Second, distillation from complementary teachers can further improve performance, with dual-teacher GLACIER variants (Mi-Mo) in some cases surpassing single-teacher models, suggesting that integrating knowledge from multiple teachers can yield additional gains.

Beyond predictive performance, practical deployment requires models to be computationally efficient. We therefore compared model performance (AUROC for classification and RMSE regression tasks) against parameter count and inference latency. Details on the experimental setup are provided in Appendix \ref{appendix:latency_exp_details}. We visualize model performance compared to parameter count and latency in Figures \ref{fig:teaser} and  \ref{fig:teaser_RMSE}. Notably, we show that GLACIER achieves high AUROC and RMSE with a significantly smaller parameter count, outperforming large baseline models such as ChemFM (1 billion parameters). Moreover, GLACIER demonstrates superior performance while maintaining more efficient inference latency over other models. These insights can be leveraged for training a smaller, faster GLACIER model from a strong teacher model such as MiniMol or MolFormer. 

\subsection{RQ3: Embedding interpretability}
To evaluate whether GLACIER learns interpretable molecular representations, we randomly sampled $1,000$ molecules from the Enamine REAL database that were not included in the pretraining set. These molecules were embedded using a GLACIER model pretrained with a single MiniMol teacher. For each molecular pair, we compared structural similarity, measured by the Tanimoto coefficient between Morgan2 fingerprints \cite{tanimotoRef}, with representation similarity, measured as the cosine similarity between their GLACIER embeddings. The resulting Pearson correlation ($r=0.48$; Figure \ref{fig:embedding_vis}) indicates that GLACIER preserves topological similarity in the learned representation space. In particular, structurally similar molecules tend to be embedded closer together, while the representations still capture information beyond that encoded by conventional molecular fingerprints.

Beyond assessing the latent-space organization of chemical structures, we further examined whether molecules with similar properties are mapped to nearby regions in the representation space \cite{vanDerMaaten2008tSNE}. To this end, we projected the 512-dimensional GLACIER embeddings of molecules from the MoleculeNet ESOL dataset into a two-dimensional t-SNE space for visualization. The resulting projection reveals clear clusters of chemically related compounds.

Together, these findings suggest that GLACIER learns structured, property-aware representations that are well suited for transfer to diverse downstream molecular prediction tasks.

\begin{table}[t]
\caption{Ablation study comparing Concatenation vs Finsler fusion using RMSE scores on MoleculeNet Best results are marked in \textbf{bold}, and the second-best results are \underline{underlined}. $\downarrow$: lower is better. Values represent means and their standard deviations from three independent runs.}
\label{tab:reg_ablation}
\centering 
\small
\begin{tabular}{lccc}
\toprule
Method & ESOL $\downarrow$ & LIPO $\downarrow$ & Avg $\downarrow$ \\ \midrule

\multicolumn{4}{l}{\textit{MolFormer Teacher}} \\
GLACIER (Concat) & \underline{0.968\textsubscript{$\pm$ 0.201}} & \textbf{0.858\textsubscript{$\pm$ 0.029}} & \textbf{0.913} \\
GLACIER (Attention) & 1.051\textsubscript{$\pm$ 0.177} & 0.917\textsubscript{$\pm$ 0.024} & 0.984 \\
GLACIER (Finsler) & \textbf{0.939\textsubscript{$\pm$ 0.105}} & \underline{0.913\textsubscript{$\pm$ 0.015}} & \underline{0.926} \\

\midrule

\multicolumn{4}{l}{\textit{MiniMol Teacher}} \\
GLACIER (Concat) & \underline{0.919\textsubscript{$\pm$ 0.172}} & \underline{0.805\textsubscript{$\pm$ 0.013}} & \underline{0.862} \\
GLACIER (Attention) & 1.287\textsubscript{$\pm$ 0.379} & 0.823\textsubscript{$\pm$ 0.020} & 1.055 \\
GLACIER (Finsler) & \textbf{0.831\textsubscript{$\pm$ 0.071}} & \textbf{0.780\textsubscript{$\pm$ 0.025}} & \textbf{0.806} \\

\bottomrule
\end{tabular}
\end{table}

\begin{table}[t]
\caption{Average performance across all classification and regression tasks in the modality ablation study using GLACIER with MiniMol as a teacher. Best results are marked in \textbf{bold}, and second-best are \underline{underlined}. $\uparrow$: higher is better; $\downarrow$: lower is better.}
\label{tab:pair_single_modality_ablation}
\centering
\small
\begin{tabular}{ccc|cc}
\toprule
\multicolumn{3}{c|}{Modality} & \multicolumn{2}{c}{Performance} \\ 
\cmidrule(r){1-3} \cmidrule(l){4-5}
Graph & Text & Tabular & Avg AUROC $\uparrow$ & Avg RMSE $\downarrow$ \\ 
\midrule

\checkmark & \checkmark & \checkmark & \textbf{0.799} & \textbf{0.806} \\

\midrule
\checkmark &  $\times$  & \checkmark & \underline{0.793} & \underline{0.828} \\
\checkmark & \checkmark & $\times$ & 0.792 & 0.942 \\
$\times$ & \checkmark & \checkmark & 0.777 & 0.890 \\

\midrule
\checkmark & $\times$ & $\times$ & 0.781 & 1.011 \\
$\times$ & \checkmark & $\times$ & 0.769 & 1.129 \\
$\times$ & $\times$ & \checkmark & 0.760 & 1.023 \\

\bottomrule
\end{tabular}
\end{table}

\subsection{RQ4: Ablation studies} 
We conducted a series of three ablation studies designed to isolate the contributions of individual model components. First, we evaluated the proposed Finsler fusion mechanism against two widely used multimodal integration strategies, concatenation and cross-attention, using both MolFormer and MiniMol as teacher models. As shown in Tables \ref{tab:cls_ablation} and \ref{tab:reg_ablation}, Finsler fusion provided a modest but consistent improvement over both baselines across classification and regression benchmarks. When MolFormer is used as the teacher model, the three fusion strategies achieve comparable performance, suggesting that the benefits of more sophisticated fusion are limited in this setting. In contrast, with MiniMol as the teacher, Finsler fusion substantially outperforms standard cross-attention, yielding higher average performance on both classification (AUROC: 0.799 vs. 0.783) and regression (RMSE: 0.806 vs. 1.055) tasks. These results indicate that the effectiveness of multimodal fusion is influenced by the choice of teacher model, with Finsler fusion providing the greatest benefit when paired with stronger teachers and demonstrating its potential to further enhance distilled molecular representations.

Second, to assess the contribution of each modality, we compared the full trimodal GLACIER model against both pairwise bimodal (graph+text, graph+tabular, and text+tabular) and unimodal (graph, text, and tabular) variants. As shown in Table \ref{tab:pair_single_modality_ablation}, the full model with MiniMol as the teacher consistently outperforms all reduced-modality configurations, highlighting the complementary nature of the three modalities and their joint contribution to more robust and generalizable representations. Detailed performance tables are provided in Tables \ref{tab:modality_ablation_class} and \ref{tab:modality_ablation_reg} in Appendix \ref{appendix:modality_abl}. 

Third, to examine the impact of pretraining scale, we evaluated GLACIER pretrained on datasets of varying sizes (10,000, 50,000, 100,000, and 500,000 randomly sampled molecules) using MiniMol as the teacher model. As shown in Table \ref{tab:scaling_analysis}, performance improves rapidly with increasing data size, demonstrating the data efficiency of the distillation framework, which already achieves strong results with only 10,000 molecules. Gains then plateau, with performance peaking around 100,000 molecules and remaining stable or slightly decreasing at larger scales. This behavior is consistent with the compact capacity of the student model (approximately 5\% of the parameters of larger foundation models) and the nature of the distillation objective, which can saturate once sufficient coverage of the teacher’s knowledge is achieved. Similar scaling patterns have been reported in prior work \cite{kaplan2020scaling, nakkiran2021deep}. Additional experimental results are provided in Tables \ref{tab:data_scaling_classification} and \ref{tab:data_scaling_regression} in Appendix \ref{appendix:scaling_analysis} and Table \ref{tab:finetune_esol_performance} in Appendix \ref{appendix:finetune}.

\begin{table}[t]
\caption{Average performance across all classification and regression tasks at different pretraining dataset sizes using GLACIER with MiniMol as a teacher. The best results are marked in \textbf{bold}, and the second-best results are \underline{underlined}. $\uparrow$: higher is better; $\downarrow$: lower is better.}
\label{tab:scaling_analysis}
\centering
\small
\begin{tabular}{lcc}
\toprule
Pretraining Size & Avg AUROC $\uparrow$ & Avg RMSE $\downarrow$ \\ 
\midrule

500K & \underline{0.798} & \underline{0.846} \\
100K & \textbf{0.799} & \textbf{0.806} \\
50K & 0.779 & 0.995 \\
10K & 0.775 & 0.979 \\
\bottomrule
\end{tabular}
\end{table}

\section{Conclusions}
In this paper, we present GLACIER, a multimodal foundation model that distills complementary knowledge from large teacher models via contrastive learning. GLACIER introduces a Finsler geometry-aware fusion mechanism that bridges asymmetric modality gaps through learnable drift and dynamic gating, enabling effective integration of graph, text, and tabular modalities.

Despite being pretrained on only 100,000 drug-like compounds, GLACIER achieves strong and consistent performance across 11 molecular benchmark datasets while maintaining high inference efficiency, demonstrating that compact multimodal models can rival larger and more resource-intensive approaches.

More broadly, this work highlights the promise of multimodal distillation frameworks for scalable molecular learning and efficient discovery of compounds with desirable properties. In large-scale virtual screening settings involving billions of candidates, even modest improvements in predictive accuracy can substantially influence the ranking of top-scoring molecules and downstream experimental prioritization. By integrating complementary chemical information into a unified representation space, GLACIER supports a wide range of molecular discovery pipelines, including virtual screening and lead optimization. To facilitate further research and adoption, we release our code and models at \href{https://github.com/eemokey/glacier}{https://github.com/eemokey/glacier}.

\section{Limitations and ethical considerations}
The results presented here suggest that GLACIER can efficiently distill knowledge from large teacher models into a compact multimodal representation while retaining strong predictive performance across diverse downstream tasks. Nevertheless, three caveats are worth noting.

First, GLACIER relies on the availability of strong teachers and therefore cannot be considered a fully standalone foundation model. Although knowledge from multiple teachers can be distilled into a single student, our current implementation does not consistently improve upon the strongest teacher and may instead converge toward their average performance. Future work may explore more effective strategies to combine complementary knowledge derived from multiple teachers. 

Second, unlike conventional Euclidean attention mechanisms, the proposed fusion module inherits the complexities of asymmetry in Finsler geometry, such as the parameters of the Finsler fusion module do not admit a closed-form solution and may converge to local minima during optimization \cite{dages2025finsler}.

Third, as with many models developed for molecular property prediction, there is potential for misuse. Models trained on biological and toxicity-related data could, in principle, be applied to the design of harmful compounds. Responsible deployment and appropriate safeguards are therefore important considerations for future applications of this work.

However, these limitations should not obscure the central finding of this study: a compact and nimble multimodal student model that achieves performance competitive with substantially larger foundation models. The results suggest that knowledge distillation offers a promising path toward efficient and deployable molecular learning systems.

\begin{acks}
E.N. was supported by NSF GRFP (DGE-1842487). A.L. was supported by the SciLifeLab \& Wallenberg Data Driven Life Science (DDLS) Program (grant: KAW 2020.0239), the Swedish Research Council (VR grant 2025-06662), and the Laboratory for Molecular Infection Medicine Sweden (MIMS) (KAW 2023.0159). This research was enabled by resources provided by the National Academic Infrastructure for Supercomputing in Sweden (NAISS), partially funded by the Swedish Research Council through grant agreement no. 2022-06725. A.L. thanks OpenEye Scientific Software for the use of OEToolkits at no cost. The authors thank Grace Yin for the artistic illustration of the GLACIER icon and Elizabeth Fife, Defu Cao, Robert Winn, Mike Gee, Bryce Kan, and Chong Liu for their feedback on the manuscript. 
\end{acks}

\section*{GenAI disclosure}
Gemini and ChatGPT were used to refine writing grammar and construct minor code snippets. All outputs were reviewed and verified by the authors prior to inclusion.

\bibliographystyle{unsrtnat} 
\bibliography{references}

\clearpage 
\onecolumn 
\appendix

\section{Implementation details}
\label{appendix:implementation_details}

\subsection{Model configuration}
The details of GLACIER's architecture are presented in Table \ref{tab:model_config}.

\begin{table}[h]
\centering
\caption{GLACIER Model Architecture}
\label{tab:model_config}
\small
\begin{tabular}{llc}
\hline
\textbf{Component} & \textbf{Subcomponent} & \textbf{Configuration} \\
\hline
\textit{Graph Encoder} & Message Passing Steps ($K$) & 3 \\
 & Output Dimension & 300 \\
 & Readout Mechanism & Attentive Aggregation \\
\hline
\textit{Text Encoder} & Transformer Layers ($N$) & 2 \\
& Heads & 8 \\
 & Hidden Dimension ($d_{text}$) & 128 \\
 & Max Sequence Length ($L$) & 512 \\
 & BPE Vocabulary Size ($V$) & 8,000 \\
\hline
\textit{Tabular Encoder} & Input Feature Dimension & 217 \\
\hline
\textit{Fusion} 
 & Modality Projections & 3-layer MLP \\
 & Geometry Parameters & $\alpha$, $\lambda$, $\boldsymbol{\omega}$ \\
\hline
\textit{Distillation} 
 & Teacher Projections & 2-layer MLP \\
 & Internal Activations & GELU \\
\end{tabular}
\end{table}

\subsection{Training dynamics and hardware}
We optimized the network using AdamW with a uniform weight decay of 0.01 across all modules and a cosine learning rate scheduler with warmup. Module-specific learning rates were set to $3 \times 10^{-4}$ for the text encoder and $1 \times 10^{-3}$ for the graph encoder, tabular encoder, and fusion components. To prevent overfitting and encourage robust multimodal learning, we applied a dropout of 0.1 and and SMILES data augmentation \cite{Bjerrum2017SMILESEA}. Pretraining and downstream inference was performed with a batch size of 1024 on a workstation equipped with an Intel Core i9-13900HX processor, 32GB of system RAM, and a single NVIDIA GeForce RTX 4080 GPU (12GB VRAM). Details on the pretraining setup and hardware are presented in Table \ref{tab:training_dynamics}. 

\begin{table}[h]
\centering
\caption{GLACIER Training Dynamics and Hardware}
\label{tab:training_dynamics}
\small
\begin{tabular}{llc}
\hline
\textbf{Component} & \textbf{Parameter} & \textbf{Configuration} \\
\hline
\textit{Optimization} & Optimizer & AdamW \\
 & Scheduler & Cosine with Warmup \\
 & Batch Size & 1024 \\
 & Max Epochs & 250 \\
 & Weight Decay & 0.01 \\
 & Graph / Tabular / Fusion LR & $1 \times 10^{-3}$ \\
 & Text LR & $3 \times 10^{-4}$ \\
\hline
\textit{Loss} & InfoNCE Temperature ($\tau$) & 0.07 \\
 & Min. Contribution Floor ($\epsilon$) & 0.1 \\
\hline
\textit{Regularization} & Fusion Modality Dropout & 0.1 \\
 & Augmentation & SMILES Canonicalization \\
\hline
\textit{Hardware} & Local Hardware & 1x RTX 4080 GPU \\
 & NAISS Hardware \cite{naiss_softwarecite} &  NVIDIA Tesla T4 GPU \\
\hline
\end{tabular}
\end{table}

\subsection{Similarity between training datasets}
\label{appendix:pretraining_sim}
\begin{figure}[h]
  \includegraphics[width=0.6\linewidth]{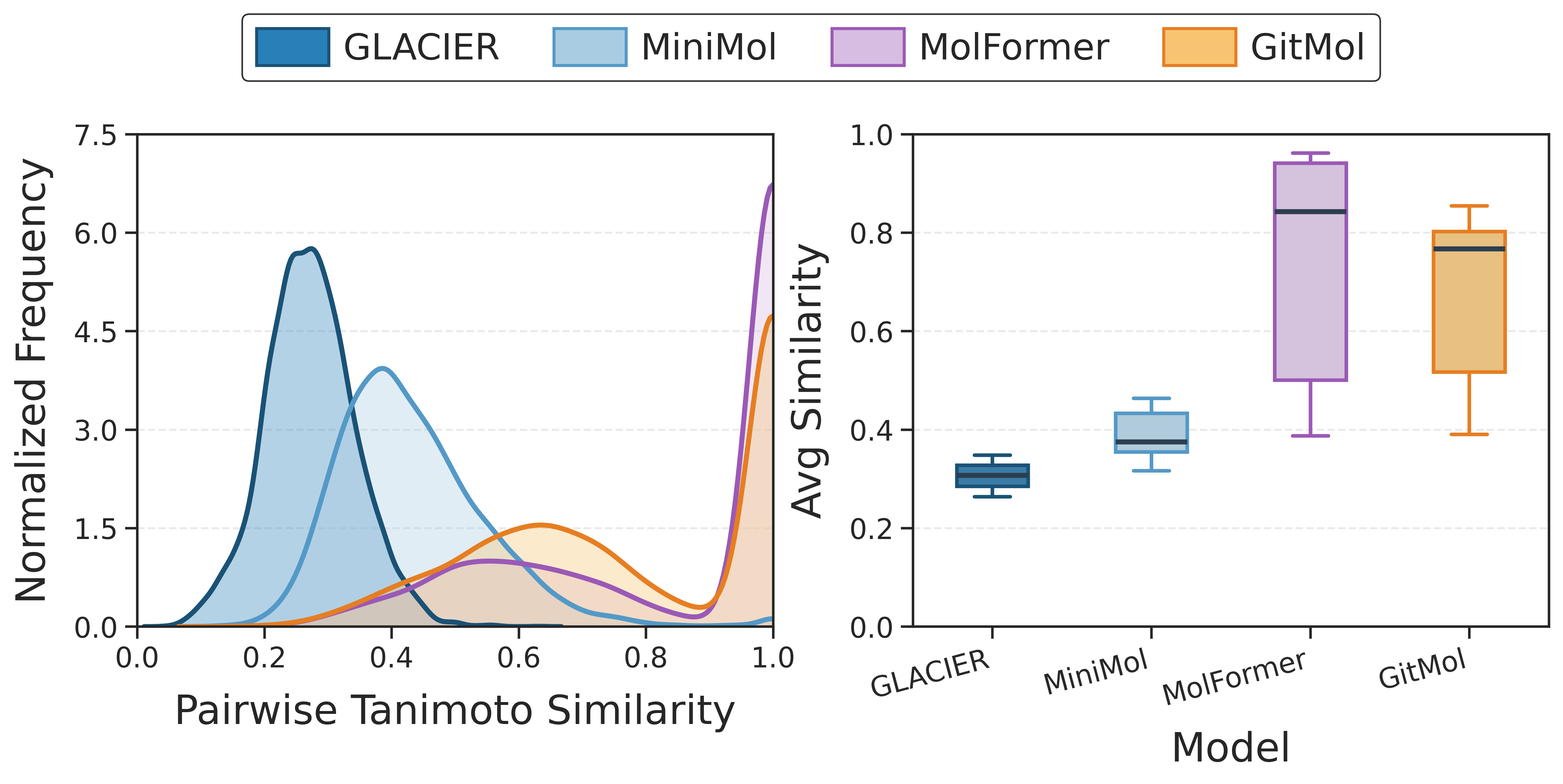}
  \caption{Distribution of Tanimoto similarity scores between pretraining and benchmark datasets. (Left) Nearest-neighbor Tanimoto similarity distribution for the AMES dataset. (Right) Distribution of dataset-wide average Tanimoto similarity scores across all 11 evaluation benchmarks. Horizontal lines within the boxes denote the median value, while the outer boundaries outline the interquartile range (IQR).}
  \Description{Distribution plot of similarity between models.}
 \label{fig:tanimoto_sim}
\end{figure}

To assess the degree of structural overlap between pretraining and downstream datasets, we measured the similarity between benchmark molecules and molecules from each model's pretraining corpus. For GLACIER, we used all 100,000 pretraining molecules, while for models with publicly available pretraining data (e.g., Git-Mol and MolFormer), we randomly sampled 100,000 molecules. For each benchmark molecule, we computed the maximum nearest-neighbor Tanimoto similarity to any molecule in the corresponding pretraining subset using Morgan fingerprints (radius = 2, 1024 bits) generated with RDKit (version 2025.09.3), where Tanimoto similarity corresponds to the Jaccard index between fingerprint bit vectors \cite{tanimotoRef}. We then averaged these maximum similarities across each benchmark dataset to quantify its structural overlap with the pretraining corpus. As shown in Figure \ref{fig:tanimoto_sim}, GLACIER was pretrained on molecules that are structurally distinct from those in the downstream benchmarks, with average maximum Tanimoto similarities of at most 0.35. While indirect exposure through the teacher models cannot be ruled out, these results suggest minimal direct overlap between GLACIER's pretraining data and the evaluation datasets. In contrast, Git-Mol and MolFormer exhibit substantially higher overlap, with average maximum similarities exceeding 0.70 on 7 of the 11 benchmarks. This indicates that molecules in their pretraining corpora are often highly similar to those in downstream datasets, potentially conferring an advantage during transfer learning.

\subsection{Description of tabular data}
\label{appendix:rdkit_desc}
We used the 217 descriptors as computed by the RDKit (version 2025.09.3) \cite{rdkitcite}. These descriptors include molecular properties such as molecular weight, logP, and the number of hydrogen bond donors and acceptors, as described in Table \ref{tab:rdkit_descriptors}.

\begin{table*}[h]
\centering
\caption{Physicochemical descriptors used as the tabular modality input
$\mathbf{x}_{\text{tab}} \in \mathbb{R}^{217}$ in GLACIER, computed by RDKit. 
}
\label{tab:rdkit_descriptors}
\small
\setlength{\tabcolsep}{2pt}
\begin{tabular}{p{2cm}|c|p{14.5cm}}
\toprule
\textbf{Category} & \textbf{Count} & \textbf{Descriptors} \\
\midrule

Physicochemical \newline properties
& 16
& \texttt{SPS}, \texttt{MolWt}, \texttt{HeavyAtomMolWt}, \texttt{ExactMolWt}, \texttt{MaxPartialCharge}, \texttt{MinPartialCharge}, \texttt{MaxAbsPartialCharge}, \texttt{MinAbsPartialCharge}, \texttt{FpDensityMorgan1}, \texttt{FpDensityMorgan2}, \texttt{FpDensityMorgan3}, \texttt{LabuteASA}, \texttt{TPSA}, \texttt{Phi}, \texttt{MolLogP}, \texttt{MolMR}
\\

\midrule

Drug-likeness
& 27
& \texttt{qed}, \texttt{NumValenceElectrons}, \texttt{NumRadicalElectrons}, \texttt{FractionCSP3}, \texttt{HeavyAtomCount}, \texttt{NHOHCount}, \texttt{NOCount}, \texttt{NumAliphaticCarbocycles}, \texttt{NumAliphaticHeterocycles}, \texttt{NumAliphaticRings}, \texttt{NumAmideBonds}, \texttt{NumAromaticCarbocycles}, \texttt{NumAromaticHeterocycles}, \texttt{NumAromaticRings}, \texttt{NumAtomStereoCenters}, \texttt{NumBridgeheadAtoms}, \texttt{NumHAcceptors}, \texttt{NumHDonors}, \texttt{NumHeteroatoms}, \texttt{NumHeterocycles}, \texttt{NumRotatableBonds}, \texttt{NumSaturatedCarbocycles}, \texttt{NumSaturatedHeterocycles}, \texttt{NumSaturatedRings}, \texttt{NumSpiroAtoms}, \texttt{NumUnspecifiedAtomStereoCenters}, \texttt{RingCount}

\\

\midrule

Topological \&\newline structural
& 20
& \texttt{AvgIpc}, \texttt{BalabanJ}, \texttt{BertzCT}, \texttt{Chi0}, \texttt{Chi0n}, \texttt{Chi0v}, \texttt{Chi1}, \texttt{Chi1n}, \texttt{Chi1v}, \texttt{Chi2n}, \texttt{Chi2v}, \texttt{Chi3n}, \texttt{Chi3v}, \texttt{Chi4n}, \texttt{Chi4v}, \texttt{HallKierAlpha}, \texttt{Ipc}, \texttt{Kappa1}, \texttt{Kappa2}, \texttt{Kappa3}
\\
\midrule

VSA surface area bins
& 36
& \texttt{PEOE\_VSA1}, \texttt{PEOE\_VSA10}, \texttt{PEOE\_VSA11}, \texttt{PEOE\_VSA12}, \texttt{PEOE\_VSA13}, \texttt{PEOE\_VSA14}, \texttt{PEOE\_VSA2}, \texttt{PEOE\_VSA3}, \texttt{PEOE\_VSA4}, \texttt{PEOE\_VSA5}, \texttt{PEOE\_VSA6}, \texttt{PEOE\_VSA7}, \texttt{PEOE\_VSA8}, \texttt{PEOE\_VSA9}, \texttt{SMR\_VSA1}, \texttt{SMR\_VSA10}, \texttt{SMR\_VSA2}, \texttt{SMR\_VSA3}, \texttt{SMR\_VSA4}, \texttt{SMR\_VSA5}, \texttt{SMR\_VSA6}, \texttt{SMR\_VSA7}, \texttt{SMR\_VSA8}, \texttt{SMR\_VSA9}, \texttt{SlogP\_VSA1}, \texttt{SlogP\_VSA10}, \texttt{SlogP\_VSA11}, \texttt{SlogP\_VSA12}, \texttt{SlogP\_VSA2}, \texttt{SlogP\_VSA3}, \texttt{SlogP\_VSA4}, \texttt{SlogP\_VSA5}, \texttt{SlogP\_VSA6}, \texttt{SlogP\_VSA7}, \texttt{SlogP\_VSA8}, \texttt{SlogP\_VSA9}
\\
\midrule

EState indices
& 25
& \texttt{MaxAbsEStateIndex}, \texttt{MaxEStateIndex}, \texttt{MinAbsEStateIndex}, \texttt{MinEStateIndex}, \texttt{EState\_VSA1}, \texttt{EState\_VSA10}, \texttt{EState\_VSA11}, \texttt{EState\_VSA2}, \texttt{EState\_VSA3}, \texttt{EState\_VSA4}, \texttt{EState\_VSA5}, \texttt{EState\_VSA6}, \texttt{EState\_VSA7}, \texttt{EState\_VSA8}, \texttt{EState\_VSA9}, \texttt{VSA\_EState1}, \texttt{VSA\_EState10}, \texttt{VSA\_EState2}, \texttt{VSA\_EState3}, \texttt{VSA\_EState4}, \texttt{VSA\_EState5}, \texttt{VSA\_EState6}, \texttt{VSA\_EState7}, \texttt{VSA\_EState8}, \texttt{VSA\_EState9}
\\
\midrule

BCUT2D
& 8
& \texttt{BCUT2D\_MWHI}, \texttt{BCUT2D\_MWLOW}, \texttt{BCUT2D\_CHGHI}, \texttt{BCUT2D\_CHGLO}, \texttt{BCUT2D\_LOGPHI}, \texttt{BCUT2D\_LOGPLOW}, \texttt{BCUT2D\_MRHI}, \texttt{BCUT2D\_MRLOW}
\\
\midrule

Fragment\newline descriptors
& 85
& \texttt{fr\_Al\_COO}, \texttt{fr\_Al\_OH}, \texttt{fr\_Al\_OH\_noTert}, \texttt{fr\_ArN}, \texttt{fr\_Ar\_COO}, \texttt{fr\_Ar\_N}, \texttt{fr\_Ar\_NH}, and 77 additional descriptors (full list: \texttt{rdkit.Chem.Fragments}) \\
\bottomrule
\end{tabular}
\end{table*}

\newpage

\section{Evaluation tasks}
\label{appendix:evaluation_task}
We evaluated the proposed GLACIER framework in various classification and regression tasks to assess its performance and applicability scope. 

\subsection{Classification metrics}
To quantify the models' performance on binary and multi-label classification tasks, we utilized the Area Under the ROC Curve (AUROC) metric, which measures the discriminative ability and is calculated as the area under the True Positive Rate (TPR) versus False Positive Rate (FPR) curve:

\begin{equation}
    \text{AUROC} = \int_{0}^{1} \text{TPR}(\text{FPR}^{-1}(t)) \, dt
\end{equation}

\subsection{Regression metrics}
To quantify the models' performance on regression property prediction tasks, we utilized the Root Mean Squared Error (RMSE) metric, which measures the square root of the average squared differences between predicted and actual values, heavily penalizing larger errors:
\begin{equation}
    \text{RMSE} = \sqrt{\frac{1}{N} \sum_{i=1}^{N} (y_i - \hat{y}_i)^2}
\end{equation}
In which $y_i$ is the ground truth, $\hat{y}_i$ is the predicted value, and $\bar{y}$ is the mean of the ground truth values for $N$ samples.

\subsection{Robustness metric}
To assess the models' performance consistency, we report empirical means with their corresponding standard deviations (STDEV):
\begin{equation}
    \text{STDEV} = \sqrt{\frac{1}{S - 1} \sum_{s=1}^{S} (m_s - \bar{m})^2}
\end{equation}
where $S=3$ is the total number of scaffold splits, $m_s$ represents the evaluation metric result for the $s$-th split, and $\bar{m}$ denotes the mean metric across all splits.

\subsection{Scaffold splits}
To provide a more realistic assessment of model generalization to unseen chemical structures, we evaluated all methods using scaffold-based data splits \cite{wu2018moleculenet}. For a fair comparison, all models are trained and evaluated using identical scaffold splits and random seeds. Because scaffold splitting enforces structural dissimilarity between training and test molecules, it is substantially more challenging than random splitting and can introduce considerable performance variability, particularly on smaller datasets where the number of unique scaffolds is limited. In this context, we observed high variance for certain baselines (e.g., 0.787 ± 0.121 AUROC for MiniMol on the E-Sub dataset). Importantly, elevated standard deviations are not observed consistently across all models or datasets, suggesting that this effect is dataset- and model-dependent.

\subsection{Benchmark dataset details}
\label{appendix:dataset_details}
A numerical overview of the benchmark datasets is provided in Table \ref{tab:dataset_stats}. Descriptions of each dataset are provided in Table \ref{tab:benchmark_description}. 

\begin{table}[b]
\caption{Benchmark dataset statistics with molecular counts and class distribution.}
\label{tab:dataset_stats}
\small
\begin{tabular}{c|c|c|c|c}
\toprule
\textbf{Dataset} & \textbf{Benchmark} & \textbf{Task} & \textbf{\# Cmpds} & \textbf{Positive \%} \\
\midrule
AMES & TDC & Class & 7,255 & 54.46 \\
BBB  & TDC & Class & 1,972 & 76.01 \\
E-Sub & TDC & Class & 664 & 28.77 \\
E-Inh & TDC & Class & 13,104 & 19.13 \\
PAMPA & TDC & Class & 2,034 & 85.50 \\
Pgp & TDC & Class & 1,212 & 53.38 \\
hERG & TDC & Class & 13,192 & 49.89 \\
Tox21 & MoleculeNet & Class & 7,730 & 3.93 \\
ToxCast & MoleculeNet & Class & 8,250 & 5.19 \\
\hline
ESOL & MoleculeNet & Regr & 1,117 & -- \\
LIPO & MoleculeNet & Regr & 4,200 & -- \\
\bottomrule
\end{tabular}
\end{table}

\begin{table*}[t]
\caption{Benchmark MoleculeNet and TDC Datasets Details}
\label{tab:benchmark_description}
\small
\setlength{\tabcolsep}{3pt}
\begin{tabular}{|p{2cm}|l|p{4.5cm}|p{9cm}|}
\toprule
\textbf{Benchmark} & \textbf{Dataset} & \textbf{Prediction Task} & \textbf{Task Description} \\
\midrule
\multirow{7}{2.2cm}{TDC \cite{wu2018moleculenet}} & AMES  & Mutagenicity Prediction & Binary classification. Predict whether a compound can cause DNA damage or mutations. \\
\cline{2-4}
& BBB & Blood-Brain Barrier Penetration & Binary classification. Predict whether a compound can cross the blood-brain barrier\\
\cline{2-4} 
& E-Sub  & CYP2D6 Enzyme Substrate Prediction & Binary classification. Predict whether a compound is a substrate of the metabolic CYP2D6 enzyme\\
\cline{2-4} 
& E-Inh  & CYP2D6 Enzyme Inhibition Prediction & Binary classification. Predict whether a compound is an inhibitor of the metabolic CYP2D6 enzyme\\
\cline{2-4} 
& PAMPA  & Parallel Artificial Membrane Permeability Assay (PAMPA) Prediction & Binary classification. Predict whether a compound has high permeability (1) or low-to-moderate permeability (0) in PAMPA assay\\
\cline{2-4} 
& Pgp  & P-glycoprotein Inhibition (Pgp) Inhibition Prediction & Binary classification. Predict whether a compound is a Pgp inhibitor\\
\cline{2-4} 
& hERG  & Human Ether-a-go-go-Related Gene (hERG) Cardiotoxicity Prediction & Binary classification. Predict whether a compound blocks (1, <10$\mu$M) or not blocks (0, $\geq$10$\mu$M)\\

\midrule
\multirow{7}{2.2cm}{MoleculeNet \cite{huang2021therapeutics}} & Tox21 & Toxicity Assessment & Multi-label classification. Predict whether a compound is toxic across 12 different toxicity pathways\\
\cline{2-4} 
& ToxCast  & Toxicity Prediction & Multi-label classification. Predict whether a compound is toxic across 617 biological assays\\
\cline{2-4} 
& ESOL  & Water Solubility Prediction & Regression. Predict the aqueous solubility of a compound\\
\cline{2-4} 
& LIPO  & Lipophilicity Prediction & Regression. Predict how well a compound dissolves in lipid environments\\
\bottomrule
\end{tabular}
\end{table*}

\section{Additional results}
\label{appendix: additonal_results}

\subsection{Selection of teacher models}
\label{appendix:teacher_selection}
Because GLACIER relies on knowledge distillation, its performance is inherently influenced by the choice of teacher models. To identify complementary teachers that provide diverse supervisory signals, we analyzed the similarity of representations produced by candidate teacher models using Centered Kernel Alignment (CKA) \cite{pmlr-v97-kornblith19a}. Specifically, we computed the linear CKA between final-layer embedding matrices generated from $100,000$ randomly selected molecules from the Enamine REAL database (65 billion, version 2024.07) \cite{enaminecite}. CKA measures the similarity between representation spaces via the normalized Hilbert-Schmidt Independence Criterion (HSIC) and is invariant to orthogonal transformations and isotropic scaling \cite{Gretton2005MeasuringSD,pmlr-v97-kornblith19a}. To maximize the diversity of distilled knowledge, we sought teacher pairs with limited representational overlap, avoiding models with highly similar embedding spaces (e.g., CKA > 0.80). Based on this analysis, we selected MiniMol and MolFormer, which exhibit a moderate CKA similarity of 0.48, indicating that they capture different aspects of molecular structure. In addition to their complementary representations, both models are well-established molecular foundation models, making them suitable choices for investigating multi-teacher distillation.

\subsection{Latency evaluation}
\label{appendix:latency_exp_details}
To ensure a fair comparison of inference efficiency, we measured the average per-molecule forward-pass latency using perf\_counter() from Python's time library on the workstation described in Table \ref{tab:training_dynamics}. Measurements were performed with a batch size of one to quantify per-molecule inference cost independent of batching effects. To isolate model execution time, data loading, tokenization, feature generation, and other preprocessing operations were excluded. For Transformer-based models, dynamic sequence padding was employed to avoid unnecessary computation on padding tokens and provide representative latency estimates.

\subsection{Scaling analysis}
\label{appendix:scaling_analysis}
The scaling analyses for individual classification and regression tasks, as well as their averages, are provided in Tables \ref{tab:data_scaling_classification} and \ref{tab:data_scaling_regression}, respectively.

\begin{table*}[h]
\caption{Effect of pretraining dataset size on classification tasks using GLACIER with MiniMol as a teacher. The best results for each dataset are marked in \textbf{bold}, and the second-best results are \underline{underlined}. $\uparrow$: the higher the better. AUROC values represent means and their standard deviations from three independent runs.}
\label{tab:data_scaling_classification}
\centering
\small
\setlength{\tabcolsep}{3pt} 
\begin{tabular}{lcccccccccc}
\toprule
Pretraining Size & AMES $\uparrow$ & BBB $\uparrow$ & Pgp $\uparrow$ & E-Sub $\uparrow$ & E-Inh $\uparrow$ & hERG $\uparrow$ & PAMPA $\uparrow$ & Tox21 $\uparrow$ & ToxCast $\uparrow$ & Avg $\uparrow$  
\\ \midrule

500K & 0.804\textsubscript{$\pm$ 0.027} & \textbf{0.902\textsubscript{$\pm$ 0.022}} & \textbf{0.920\textsubscript{$\pm$ 0.033}} & \underline{0.765\textsubscript{$\pm$ 0.091}} & \underline{0.855\textsubscript{$\pm$ 0.015}} & \textbf{0.811\textsubscript{$\pm$ 0.019}} & 0.674\textsubscript{$\pm$ 0.079} & \underline{0.785\textsubscript{$\pm$ 0.032}} & \underline{0.666\textsubscript{$\pm$ 0.007}} & \underline{0.798} \\

100K & \textbf{0.828\textsubscript{$\pm$ 0.018}} & \underline{0.895\textsubscript{$\pm$ 0.027}} & 0.900\textsubscript{$\pm$ 0.029} & \textbf{0.769\textsubscript{$\pm$ 0.121}} & \textbf{0.856\textsubscript{$\pm$ 0.011}} & \underline{0.803\textsubscript{$\pm$ 0.026}} & \textbf{0.687\textsubscript{$\pm$ 0.079}} & \textbf{0.786\textsubscript{$\pm$ 0.034}} & \textbf{0.668\textsubscript{$\pm$ 0.008}} & \textbf{0.799} \\

50K & \underline{0.812\textsubscript{$\pm$ 0.023}} & 0.854\textsubscript{$\pm$ 0.044} & 0.897\textsubscript{$\pm$ 0.040} & 0.733\textsubscript{$\pm$ 0.068} & 0.853\textsubscript{$\pm$ 0.014} & 0.772\textsubscript{$\pm$ 0.015} & \underline{0.677\textsubscript{$\pm$ 0.054}} & 0.762\textsubscript{$\pm$ 0.024} & 0.653\textsubscript{$\pm$ 0.008} & 0.779 \\

10K & 0.804\textsubscript{$\pm$ 0.023} & 0.864\textsubscript{$\pm$ 0.037} & \underline{0.902\textsubscript{$\pm$ 0.040}} & 0.726\textsubscript{$\pm$ 0.061} & 0.848\textsubscript{$\pm$ 0.011} & 0.757\textsubscript{$\pm$ 0.023} & 0.665\textsubscript{$\pm$ 0.036} & 0.765\textsubscript{$\pm$ 0.018} & 0.647\textsubscript{$\pm$ 0.013} & 0.775 \\

\bottomrule
\end{tabular}
\end{table*}

\begin{table*}[h]
\caption{Effect of pretraining dataset size on regression tasks using GLACIER with MiniMol as a teacher. The best results for each dataset are marked in \textbf{bold}, and the second-best results are \underline{underlined}. $\downarrow$: the lower the better. RMSE values represent means and their standard deviations from three independent runs.}
\label{tab:data_scaling_regression}
\centering
\small
\setlength{\tabcolsep}{3pt} 
\begin{tabular}{lccc}
\toprule
Pretraining Size & ESOL $\downarrow$ & LIPO $\downarrow$ & Avg $\downarrow$  
\\ \midrule

500K & \underline{0.955\textsubscript{$\pm$ 0.170}} & \textbf{0.737\textsubscript{$\pm$ 0.011}} & \underline{0.846} \\
100K & \textbf{0.831\textsubscript{$\pm$ 0.071}} & \underline{0.780\textsubscript{$\pm$ 0.025}} & \textbf{0.806} \\
50K & 1.152\textsubscript{$\pm$ 0.235} & 0.838\textsubscript{$\pm$ 0.006} & 0.995 \\
10K & 1.083\textsubscript{$\pm$ 0.102} & 0.875\textsubscript{$\pm$ 0.013} & 0.979 \\

\bottomrule
\end{tabular}
\end{table*} 

\subsection{Modality ablation studies}
\label{appendix:modality_abl}
To assess the contribution of each molecular representation and determine whether their integration provides complementary information, we conducted a modality ablation study comparing the full trimodal model against both pairwise bimodal and unimodal variants. The trimodal GLACIER model (MiniMol teacher) consistently achieves the best average performance on both classification and regression benchmarks, as shown in Tables \ref{tab:modality_ablation_class} and \ref{tab:modality_ablation_reg}, respectively. These results indicate that graph, text, and tabular representations capture complementary aspects of molecular structure and properties, and that their joint integration yields more robust and informative molecular representations than any subset of modalities alone.

\begin{table*}[ht]
\caption{Modality ablation study on classification tasks using GLACIER with MiniMol as a teacher. The best results for each dataset are marked in \textbf{bold}, and the second-best results are \underline{underlined}. $\uparrow$: the higher the better. AUROC values represent means and their standard deviations from three independent runs.}
\label{tab:modality_ablation_class}
\centering
\small
\setlength{\tabcolsep}{2.5pt} 
\begin{tabular}{lcccccccccc}
\toprule
Modality Configuration & AMES $\uparrow$ & BBB $\uparrow$ & Pgp $\uparrow$ & E-Sub $\uparrow$ & E-Inh $\uparrow$ & hERG $\uparrow$ & PAMPA $\uparrow$ & Tox21 $\uparrow$ & ToxCast $\uparrow$ & Avg $\uparrow$  
\\ \midrule

All Modalities & \textbf{0.828\textsubscript{$\pm$ 0.018}} & \textbf{0.895\textsubscript{$\pm$ 0.027}} & 0.900\textsubscript{$\pm$ 0.029} & \textbf{0.769\textsubscript{$\pm$ 0.121}} & \textbf{0.856\textsubscript{$\pm$ 0.011}} & \underline{0.803\textsubscript{$\pm$ 0.026}} & 0.687\textsubscript{$\pm$ 0.079} & \textbf{0.786\textsubscript{$\pm$ 0.034}} & \textbf{0.668\textsubscript{$\pm$ 0.008}} & \textbf{0.799} \\

\midrule

Graph+Tabular & \textbf{0.828\textsubscript{$\pm$ 0.020}} & 0.868\textsubscript{$\pm$ 0.028} & 0.907\textsubscript{$\pm$ 0.027} & \underline{0.760\textsubscript{$\pm$ 0.101}} & \underline{0.850\textsubscript{$\pm$ 0.011}} & 0.794\textsubscript{$\pm$ 0.029} & \underline{0.700\textsubscript{$\pm$ 0.088}} & 0.769\textsubscript{$\pm$ 0.032} & \underline{0.662\textsubscript{$\pm$ 0.011}} & \underline{0.793} \\

Graph+Text & \underline{0.823\textsubscript{$\pm$ 0.024}} & 0.868\textsubscript{$\pm$ 0.027} & \textbf{0.916\textsubscript{$\pm$ 0.028}} & 0.728\textsubscript{$\pm$ 0.069} & \textbf{0.856\textsubscript{$\pm$ 0.012}} & \textbf{0.804\textsubscript{$\pm$ 0.020}} & 0.699\textsubscript{$\pm$ 0.056} & \underline{0.774\textsubscript{$\pm$ 0.023}} & 0.659\textsubscript{$\pm$ 0.012} & 0.792 \\

Tabular+Text & 0.807\textsubscript{$\pm$ 0.027} & 0.850\textsubscript{$\pm$ 0.022} & \underline{0.914\textsubscript{$\pm$ 0.030}} & 0.731\textsubscript{$\pm$ 0.095} & 0.831\textsubscript{$\pm$ 0.009} & 0.793\textsubscript{$\pm$ 0.011} & 0.653\textsubscript{$\pm$ 0.079} & 0.762\textsubscript{$\pm$ 0.035} & 0.649\textsubscript{$\pm$ 0.003} & 0.777 \\

\midrule

Graph Only & 0.807\textsubscript{$\pm$ 0.028} & \underline{0.872\textsubscript{$\pm$ 0.040}} & 0.893\textsubscript{$\pm$ 0.025} & 0.748\textsubscript{$\pm$ 0.080} & 0.845\textsubscript{$\pm$ 0.015} & 0.795\textsubscript{$\pm$ 0.026} & 0.656\textsubscript{$\pm$ 0.108} & 0.760\textsubscript{$\pm$ 0.030} & 0.653\textsubscript{$\pm$ 0.008} & 0.781 \\

Text Only & 0.773\textsubscript{$\pm$ 0.027} & 0.833\textsubscript{$\pm$ 0.035} & 0.913\textsubscript{$\pm$ 0.029} & 0.757\textsubscript{$\pm$ 0.074} & 0.799\textsubscript{$\pm$ 0.007} & 0.744\textsubscript{$\pm$ 0.023} & \textbf{0.715\textsubscript{$\pm$ 0.081}} & 0.755\textsubscript{$\pm$ 0.032} & 0.631\textsubscript{$\pm$ 0.009} & 0.769 \\

Tabular Only & 0.751\textsubscript{$\pm$ 0.034} & 0.838\textsubscript{$\pm$ 0.009} & 0.910\textsubscript{$\pm$ 0.023} & 0.751\textsubscript{$\pm$ 0.106} & 0.800\textsubscript{$\pm$ 0.005} & 0.735\textsubscript{$\pm$ 0.021} & 0.661\textsubscript{$\pm$ 0.082} & 0.749\textsubscript{$\pm$ 0.036} & 0.650\textsubscript{$\pm$ 0.010} & 0.760 \\

\bottomrule
\end{tabular}
\end{table*}

\begin{table}[ht]
\caption{Modality ablation study on regression tasks using GLACIER with MiniMol as a teacher. The best results for each dataset are marked in \textbf{bold}, and the second-best results are \underline{underlined}. $\downarrow$: the lower the better. RMSE values represent means and their standard deviations from three independent runs.}
\label{tab:modality_ablation_reg}
\centering
\small
\setlength{\tabcolsep}{3pt} 
\begin{tabular}{lccc}
\toprule
Modality Configuration & ESOL $\downarrow$ & LIPO $\downarrow$ & Avg $\downarrow$  
\\ \midrule

All Modalities & \textbf{0.831\textsubscript{$\pm$ 0.071}} & \underline{0.780\textsubscript{$\pm$ 0.025}} & \textbf{0.806} \\

\midrule

Graph + Tabular & \underline{0.866\textsubscript{$\pm$ 0.080}} & 0.789\textsubscript{$\pm$ 0.013} & \underline{0.828} \\
Graph + Text & 1.118\textsubscript{$\pm$ 0.182} & \textbf{0.765\textsubscript{$\pm$ 0.035}} & 0.942 \\
Tabular + Text & 0.962\textsubscript{$\pm$ 0.224} & 0.819\textsubscript{$\pm$ 0.006} & 0.890 \\

\midrule

Graph Only & 1.227\textsubscript{$\pm$ 0.183} & 0.794\textsubscript{$\pm$ 0.012} & 1.011 \\
Text Only & 1.308\textsubscript{$\pm$ 0.299} & 0.949\textsubscript{$\pm$ 0.024} & 1.129 \\
Tabular Only & 1.087\textsubscript{$\pm$ 0.231} & 0.959\textsubscript{$\pm$ 0.023} & 1.023 \\

\bottomrule
\end{tabular}
\end{table}

\subsection{Model finetuning}
\label{appendix:finetune}
Throughout this work, model performance is evaluated using downstream fingerprinting, where embeddings from a single forward pass of a pretrained model are evaluated by a task-specific head. In addition to using this lightweight evaluation protocol to estimate the quality of the learned representations, we compared it against finetuning a full model end-to-end, in which all model parameters are updated. Table \ref{tab:finetune_esol_performance} shows that finetuning improves performance for both MolFormer as a teacher and GLACIER as its student model on the ESOL dataset. However, GLACIER already achieves strong performance in the downstream fingerprinting setting and remains superior after finetuning. While full finetuning provides substantial gains (1.866 to 1.108 average RMSE) for the MolFormer teacher, the relatively small improvement (0.939 to 0.882 average RMSE) observed for its GLACIER student suggests that its pretrained representations are already highly predictive. Given the increased computational cost of updating the full GLACIER model, these results support the use of a frozen backbone as an efficient and effective downstream strategy.

\begin{table}[h]
\centering
\caption{Performance comparison between downstream fingerprinting and finetuned models on the ESOL regression task. The best results are marked in \textbf{bold}. $\downarrow$: the lower the better. RMSE values are represented as means and their corresponding standard deviations from three independent runs.}
\label{tab:finetune_esol_performance}
\begin{tabular}{lcc}
\toprule
\textbf{Method} & \textbf{Downstream Fingerprinting} & \textbf{Finetuned} \\
\midrule
MolFormer & $1.866 \pm 0.553$ & $1.108 \pm 0.427$ \\
\textbf{GLACIER (MolFormer)} & $\mathbf{0.939 \pm 0.105}$ & $\mathbf{0.882 \pm 0.172}$ \\
\bottomrule
\end{tabular}
\end{table}

\section{Architecture}
\label{appendix:algs}
The architecture of the Finsler-based fusion approach is presented in Algorithm \ref{alg:finsler_fusion} and the architecture of the overall distillation pipeline is presented in Algorithm \ref{alg:training_loop}.

\begin{algorithm*}[h]
\caption{Multimodal Finsler Fusion}
\label{alg:finsler_fusion}
\begin{algorithmic}[1]
\Require $\mathbf{z}_{text}, \mathbf{z}_{graph}, \mathbf{z}_{tab} \in \mathbb{R}^d$, MLPs $\{q,k,v,drift,amp\}$, base $\lambda_{raw}$
\Ensure Fused representation $\mathbf{h}_{fused}$

\State $Q, \boldsymbol{\omega}_{raw} \gets \text{MLP}_q(\mathbf{z}_{text}),\ \text{MLP}_{drift}(\mathbf{z}_{text})$  \Comment{Get query}
\State $\boldsymbol{\omega} \gets \frac{\boldsymbol{\omega}_{raw}}{\|\boldsymbol{\omega}_{raw}\|_2 + \epsilon} \cdot \tanh(\|\boldsymbol{\omega}_{raw}\|_2)$ \Comment{Calculate drift}

\State $S \gets \{\mathbf{z}_{graph}, \mathbf{z}_{tab}\}$
\For{$\mathbf{k}_i \in S$} \Comment{Process remaining modalities}
    \State $K_i, V_i \gets \text{MLP}_k(\mathbf{k}_i),\ \text{MLP}_v(\mathbf{k}_i)$
    \State $d_i \gets \|K_i - Q\|_2 + \langle K_i - Q, \boldsymbol{\omega} \rangle$ \Comment{Calculate asymmetric distance}
\EndFor
\State $w \gets \text{Softmax}(-\{d_{graph}, d_{tab}\} / \sqrt{d})$ \Comment{Calculate attention weights}
\State $\mathbf{c} \gets \sum_{i}^{} w_i V_i$ 
\State $\alpha, \lambda \gets \text{Softplus}(\text{MLP}_{amp}(\mathbf{z}_{text})),\ \text{Softplus}(\lambda_{raw})$ \Comment{Calculate gating factor}
\State $\gamma \gets \alpha \cdot \sigma(-\min(d_i) \cdot \lambda / \sqrt{d})$ 
\State $\mathbf{\hat{z}}_{text} \gets \mathbf{z}_{text} + \gamma \mathbf{c}$ \Comment{Update text representation}
\State \Return $\text{LayerNorm}(\text{Linear}(\mathbf{z}_{graph} \parallel \mathbf{\hat{z}}_{text} \parallel \mathbf{z}_{tab}))$ 
\end{algorithmic}
\end{algorithm*}

\begin{algorithm*}[h]
\caption{Multimodal Pretraining with Student-Teacher Distillation}
\label{alg:training_loop}
\begin{algorithmic}[1]
\Require Data $\mathcal{X}$, teachers $T_{raw}$, temp $\tau$, min-trust $\epsilon$
\Ensure Dynamic distillation loss $\mathcal{L}_{total}$

\State $\mathbf{z}_{graph}, \mathbf{z}_{text}, \mathbf{z}_{tab} \gets \text{Encoders}(\mathcal{X}_{graph}, \mathcal{X}_{text}, \mathcal{X}_{tab})$ \Comment{\textit{Step 1: Feature Extraction}}

\State \Comment{\textit{Step 2: Finsler Fusion}}
\State $\mathbf{h}_{fused} \gets \text{Algorithm \ref{alg:finsler_fusion}}(\mathbf{z}_{graph}, \mathbf{z}_{text}, \mathbf{z}_{tab})$
\State $\mathbf{h}_{proj} \gets \text{Projector}_{student}(\mathbf{h}_{fused})$
\State $\mathbf{w}_{trust} \gets \sigma(\text{MLP}_{trust}(\mathbf{h}_{proj}) \cdot (1 - \epsilon) + \epsilon$

\State \Comment{\textit{Step 3: Student-Teacher InfoNCE Distillation}}
\State $\mathcal{L}_{total} \gets 0, \quad \mathbf{h}_{norm} \gets \text{Normalize}(\mathbf{h}_{proj})$ \hfill 
\Comment{Initialize loss \& normalize student}
\For{$T_i \in T_{raw}$}
   \State $T_{norm} \gets \text{Normalize}(\text{Projector}_i(T_i))$  \Comment{Project and normalize teacher}
    \State $\mathcal{L}_{NCE} \gets \text{CrossEntropy}(\mathbf{h}_{norm} T_{norm}^\top / \tau)$ \Comment{Contrastive alignment}
    \State $\mathcal{L}_{total} \gets \mathbf{w}_{\text{trust}, i} \cdot \mathcal{L}_{NCE} - \log(\mathbf{w}_{\text{trust}, i})$
\EndFor

\State \Return $\mathcal{L}_{total}$
\end{algorithmic}
\end{algorithm*}
\end{document}